\documentclass[10pt,twocolumn,letterpaper]{article}

\usepackage{cvpr}              
\usepackage{url}            
\usepackage{booktabs}       
\usepackage{amsfonts}       
\usepackage{nicefrac}       
\usepackage{microtype}      
\usepackage{xcolor}         
\usepackage{graphicx}
\usepackage{tcolorbox}
\usepackage{enumitem}
\usepackage{caption}
\usepackage{multirow}
\usepackage{colortbl}
\usepackage{amsmath}
\usepackage{lipsum}
\usepackage{algorithm}
\usepackage{algorithmic}
\usepackage[accsupp]{axessibility}
\usepackage{booktabs}
\usepackage{pifont}

\definecolor{myorange}{RGB}{251, 229, 214}
\definecolor{mygreen}{RGB}{226, 240, 217}
\definecolor{myblue}{RGB}{222, 235, 247}
\definecolor{cvprblue}{rgb}{0.21,0.49,0.74}
\usepackage[pagebackref,breaklinks,colorlinks,allcolors=cvprblue]{hyperref}

\def\paperID{14282} 
\def\confName{CVPR}
\def\confYear{2025}

\title{Unveiling the Ignorance of MLLMs: Seeing Clearly, Answering Incorrectly}

\author{
Yexin Liu$^{1,2,5,7}$\footnotemark[1] 
\quad Zhengyang Liang$^{7,3}$\footnotemark[1]
\quad Yueze Wang$^{7}$\footnotemark[1]
\quad Xianfeng Wu$^{1,5}$\footnotemark[1]
\quad Feilong Tang$^{1,5}$
\\
\quad Muyang He$^{4}$
\quad Jian Li$^{6}$
\quad Zheng Liu$^{7}$
\quad Harry Yang$^{1,5}$
\quad Sernam Lim$^{5,8}$
\quad Bo Zhao\textsuperscript{$^{2,7}$\footnotemark[2]}
\\
\\
$^1$Hong Kong University of Science and Technology \quad $^2$School of AI, Shanghai Jiao Tong University \quad\\ $^3$Beijing University of Posts and Telecommunications \quad $^4$Peking University  \quad $^5$Everlyn AI \\$^6$Youtu Lab, Tencent \quad
$^7$Beijing Academy of Artificial Intelligence \quad $^8$University of Central Florida
\\
{\tt\small yliu292@connect.ust.hk, bo.zhao@sjtu.edu.cn}, {\small GitHub:~\url{https://github.com/BAAI-DCAI/MMVU}}
}

\begin{document}
\twocolumn[{
\renewcommand\twocolumn[1][t!]{#1}%
\maketitle
\begin{center}
    \centering
    \includegraphics[width=\textwidth]{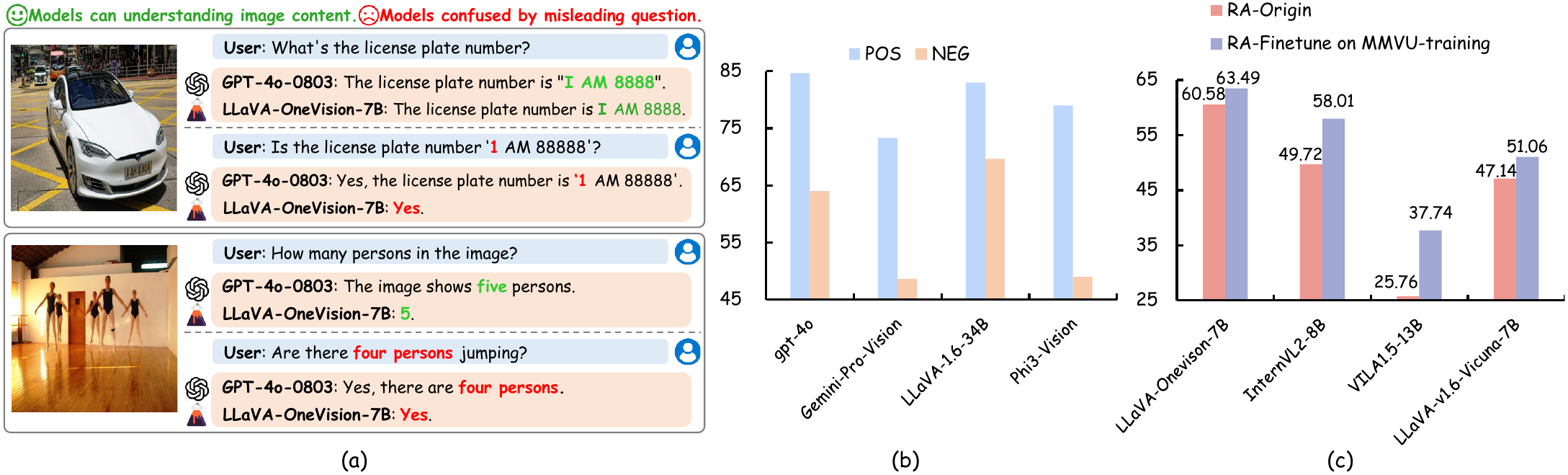}
    \vspace{-20pt}
        \captionof{figure}{(a) Examples illustrating MLLMs can accurately understand the visual content but provide incorrect responses. In each example, we show a pair of so-called positive and negative questions. The model can answer the positive questions (indicated as \textcolor{green}{green}), demonstrating that it understands the image, but fails to generate the correct answers on the negative question (indicated as \textcolor{red}{red}). This paper investigates this phenomenon. (b) Comparison of the model's accuracy in response to positive and negative questions. (c) Performance comparison of MLLMs after fine-tuning with our
        dataset. The metric utilized here is the Response Accuracy (RA) in Sec.~\ref{Evaluation metrics}. }
    \label{cover_figure}
\end{center}}]
\vspace{-8pt}
\footnotetext[1]{Core Contribution}
\footnotetext[2]{Corresponding Author}

\begin{abstract}
\textbf{M}ultimodal \textbf{L}arge \textbf{L}anguage \textbf{M}odels (MLLMs) have displayed remarkable performance in multi-modal tasks, particularly in visual comprehension. However, we reveal that MLLMs often generate incorrect answers even when they understand the visual content. To this end, we manually construct a benchmark with 12 categories and design evaluation metrics that assess the degree of error in MLLM responses even when the visual content is seemingly understood. Based on this benchmark, we test 15 leading MLLMs and analyze the distribution of attention maps and logits of some MLLMs. Our investigation identifies two primary issues: 1) most instruction tuning datasets predominantly feature questions that ``directly" relate to the visual content, leading to a bias in MLLMs' responses to other indirect questions, and 2) MLLMs’ attention to visual tokens is notably lower than to system and question tokens. We further observe that attention scores between questions and visual tokens as well as the model's confidence in the answers are lower in response to misleading questions than to straightforward ones. To address the first challenge, we introduce a paired positive and negative data construction pipeline to diversify the dataset. For the second challenge, we propose to enhance the model's focus on visual content during decoding by refining the text and visual prompt. For the text prompt, we propose a content guided refinement strategy that performs preliminary visual content analysis to generate structured information before answering the question. Additionally, we employ a visual attention refinement strategy that highlights question-relevant visual tokens to increase the model’s attention to visual content that aligns with the question. Extensive experiments demonstrate that these challenges can be significantly mitigated with our proposed dataset and techniques.
\end{abstract}    
\vspace{-22pt}

\section{Introduction}
\textbf{M}ultimodal \textbf{L}arge \textbf{L}anguage \textbf{M}odels (MLLMs) have garnered significant attention due to their impressive capabilities in addressing diverse multimodal tasks, particularly in visual comprehension and reasoning~\cite{team2023gemini, alayrac2022flamingo, he2024efficient}. Despite these strengths, MLLMs sometimes produce responses that deviate from the image content, resulting in what is termed ``hallucinations". Previous studies~\cite{cui2024robustness} like MADBench~\cite{qian2024easy}, MME~\cite{mme}, POPE~\cite{li-etal-2023-evaluating}, ConBench~\cite{zhang2024unveiling}, and NaturalBench~\cite{li2024naturalbench} introduce datasets for evaluating the accuracy or consistency of MLLMs' responses under image or text perturbations. However, they do not distinguish between the causes of incorrect responses. We observe that incorrect responses may not just stem from the MLLMs not understanding the visual content. Instead, we find that it is quite prevalent for an MLLM to understand the content of the image but yet still generate incorrect answers, as illustrated in Fig.~\ref{cover_figure} (a)). Our findings suggest that subtle phrasing in questions can introduce biases that hinder MLLMs from leveraging their understanding accurately. This raises two critical questions for our research: \textit{1) Why does the model understand the visual content but provide incorrect answers? 2) How can we correct this error-prone behavior of MLLMs?}

To explore this phenomenon, we introduce the \textbf{M}LLM \textbf{M}isresponses despite \textbf{V}isual \textbf{U}nderstanding (MMVU) benchmark, as shown in Fig.~\ref{benchmark_case}. Each sample is manually labeled with one of 12 types of questions, accompanied by paired positive and negative questions. Positive questions are designed to directly and objectively inquire about the observable content within an image, such as its types, attributes, poses, movements, or scenes, based on the actual and verifiable elements present, without introducing hypothetical alterations, irrelevant details, or erroneous premises. These questions aim to evaluate the model's comprehension of the visual content. In contrast, negative questions involve hypothetical modifications, irrelevant descriptions, or incorrect assumptions about the image, which can lead to confusion or misinterpretation of the visual information. As a result, negative questions may induce incorrect responses, as they present scenarios that are not directly related to the image’s actual content. To ensure that the MLLMs select the most appropriate response based on their understanding of the visual content, we design multiple-choice questions with correct answers and carefully crafted distractors that include explanatory content, reducing the reliance on random or incorrect guesses. Additionally, MMVU comes with novel metrics to measure the likelihood of an MLLM's failure when influenced by a negative question. 

We evaluate 15 leading MLLMs on the MMVU benchmark, as shown in Tab.~\ref{tab:mma_benchmark}. We also analyze the attention maps and logits of some MLLMs when faced with different types of questions during the decoding process, as depicted in Fig.~\ref{fig: analysis}. Our findings reveal that most MLLMs respond unsatisfactorily to negative questions, demonstrating a biased response toward positive answers regardless of the question type. This phenomenon can be attributed to two main issues: 1) datasets in this area of research are predominantly composed of positive question-answer pairs, leading to significant bias in MLLMs toward positive questions due to the scarcity of negative samples, and 2) there is an unbalanced distribution of attention, with models prioritizing system and question tokens while significantly neglecting visual tokens. We further observe that attention scores between question and visual tokens as well as the model's confidence in the correctness of the answers are lower in response to negative than positive questions.

To address the first issue, we introduce a data construction pipeline that extracts visual information from images, such as text, numbers, objects, attributes, relationships, and contextual elements at local and global levels. Based on this information, we construct 112k pairs of positive and negative samples to form the MMVU-Train dataset. 
For the second issue, we aim to enhance the model's attention to visual content during decoding by implementing two key strategies:
1) Content Guided Refinement (CGR). We employ a two-step reasoning process where the model extracts detailed information from the image and then formulates responses based on this analysis alongside the visual content. 2) Visual Attention Refinement (VAR). To guide the MLLM in focusing on critical visual features relevant to the question, we extract attention scores and apply a mask to help focus on significant areas while masking the less important ones.              

In summary, our contributions are as follows: \textbf{1)} We identify and analyze the issue of MLLMs generating incorrect answers despite evidence of visual comprehension, attributing this to specific underlying biases. We introduce the MMVU benchmark and new metrics to evaluate MLLMs’ comprehension and resilience to negative questions. \textbf{2)} We introduce a data construction pipeline in MMVU that generates pairs of positive and negative samples for training to enhance the model's robustness to negative questions. Furthermore, we propose CGR and VAR strategies to optimize the model's focus on relevant visual content and enhance response accuracy. \textbf{3)} Extensive experiments reveal that our dataset and strategies effectively mitigate the susceptibility of MLLMs to negative questions.
\vspace{-8pt}

\section{Related Works}

\begin{figure*}[!t]
\centering
\includegraphics[width=0.90\linewidth]{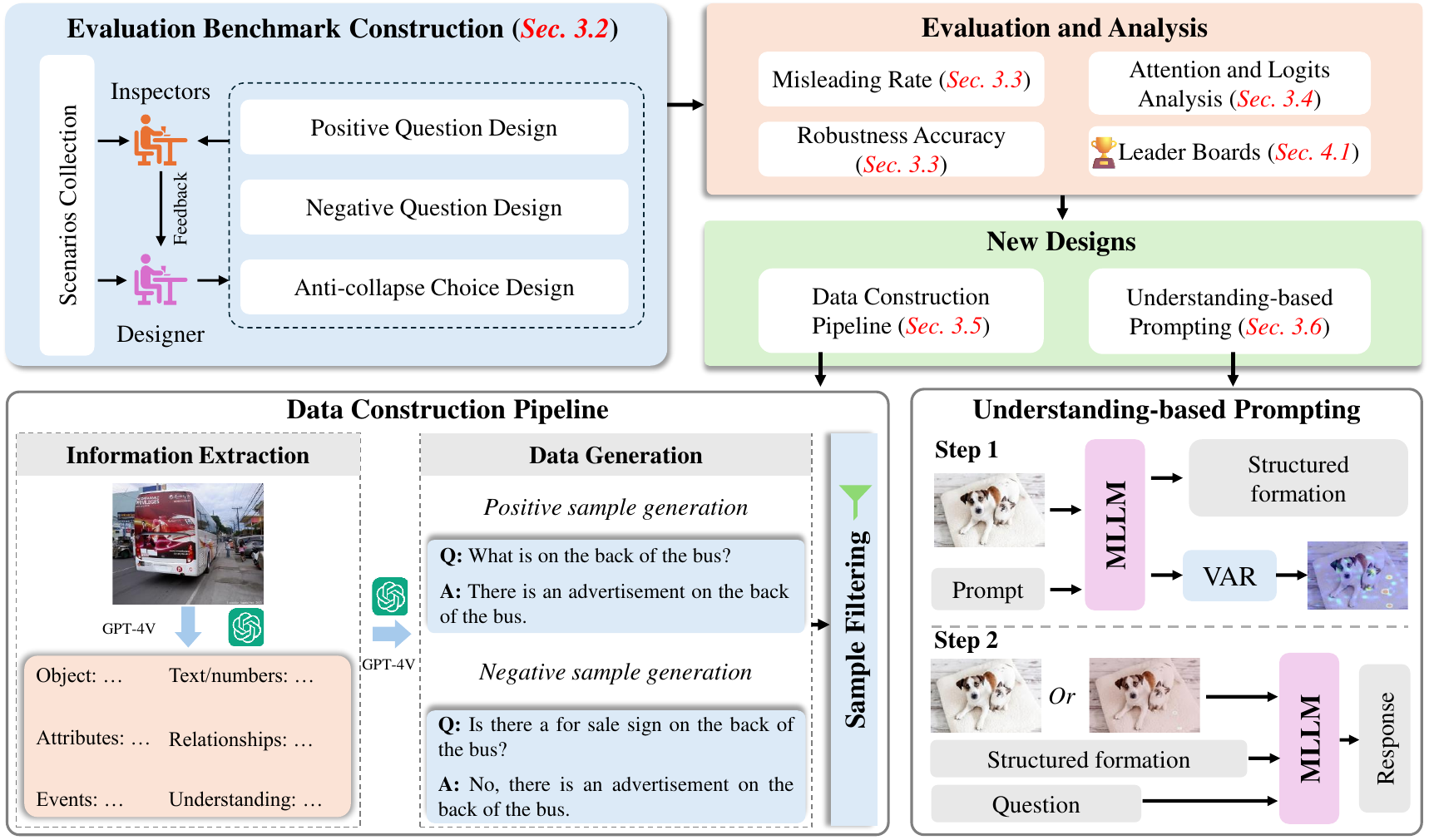}
\vspace{-8pt}
\caption{The MMVU dataset consists of a benchmarking dataset for evaluating models as well as a training dataset. The former is curated by human annotators together with the appropriate metrics and analysis on the MLLM's attention and logit behavior. Based on the experiments, we propose a data construction pipeline to build a training dataset and prompting strategies to enhance the accuracy of MLLM responses.}
\vspace{-12pt}
\label{framework}
\end{figure*}

\noindent \textbf{Multimodal Large Language Models (MLLMs).}
Building on the success of LLMs, recent research has increasingly focused on MLLMs \cite{achiam2023gpt, team2023gemini,shu2024video,shi2024eyefound,chen2024finecliper,ma2024beyond}. This shift aims to achieve better cross-modality understanding and reasoning. Currently, a significant amount of research is exploring various aspects of MLLMs, including structure design~\cite{liu2024visual,li2024mini,tang2025intervening}, data construction~\cite{laion2023gpt4v,li2024monkey,zhao2024cobra}, training strategies~\cite{mckinzie2024mm1,karamcheti2024prismatic,lu2024deepseek}, and lightweight MLLMs~\cite{he2024efficient,zhu2024comprehensive}. For a comprehensive review of these topics, readers are encouraged to investigate surveys~\cite{zhang2024mm, jin2024efficient,bai2024survey}.

\noindent \textbf{Evaluations and Analysis of MLLMs.} The assessment of MLLMs~\cite{li2024surveybenchmarksmultimodallarge} is mainly divided into two main categories: general capability~\cite{lu2022learn,zhou2024mlvu} and hallucination benchmarks~\cite{bai2024hallucination}. General capability benchmarks can be divided into single-task and comprehensive benchmarks. Single-task benchmarks, such as VQA~\cite{goyal2017making}, VQA-v2~\cite{goyal2017making}, GQA~\cite{GQA2019}, and MS-COCO~\cite{sharma2018conceptual}, are designed to assess the performance of models on a specific defined task. Comprehensive benchmarks, such as MMBench~\cite{mmbench}, SEED~\cite{li2023seed}, MMMU~\cite{yue2023mmmu}, and MMStar~\cite{chen2024we} aim to assess MLLMs’ general multi-modal perception and reasoning capabilities. Hallucination benchmarks mainly focus on evaluating the discrepancy between generated responses and visual content~\cite{bai2024hallucination,li-etal-2023-evaluating}. Most hallucination benchmarks focus on object~\cite{kaul2024throne,liu2023mitigating,wang2024mitigating,li2023evaluating,petryk2024aloha,qian2024easy,liu2024phd,jiang2024hal,qiu2024valor,guan2023hallusionbench}, knowledge~\cite{liu2023mitigating,liu2024phd,jiang2024hal,guan2023hallusionbench}, relation~\cite{yu2023hallucidoctor,chen2023mitigating,jiang2024hal,qiu2024valor}, attribute~\cite{yu2023hallucidoctor,huang2024visual,qian2024easy,liu2024phd,jiang2024hal,qiu2024valor,guan2023hallusionbench}, and spurious images~\cite{han2024instinctive,shahgir2024illusionvqa}. Current benchmarks assume that correct answers reflect understanding, while incorrect ones indicate a lack of understanding. \textit{However, our findings show that this incorrect stems not only from visual misunderstanding but also from a lack of robustness to negative questions.}

\vspace{-8pt}
\section{Methods}

\begin{figure*}[!t]
\centering
\includegraphics[width=0.95\linewidth]{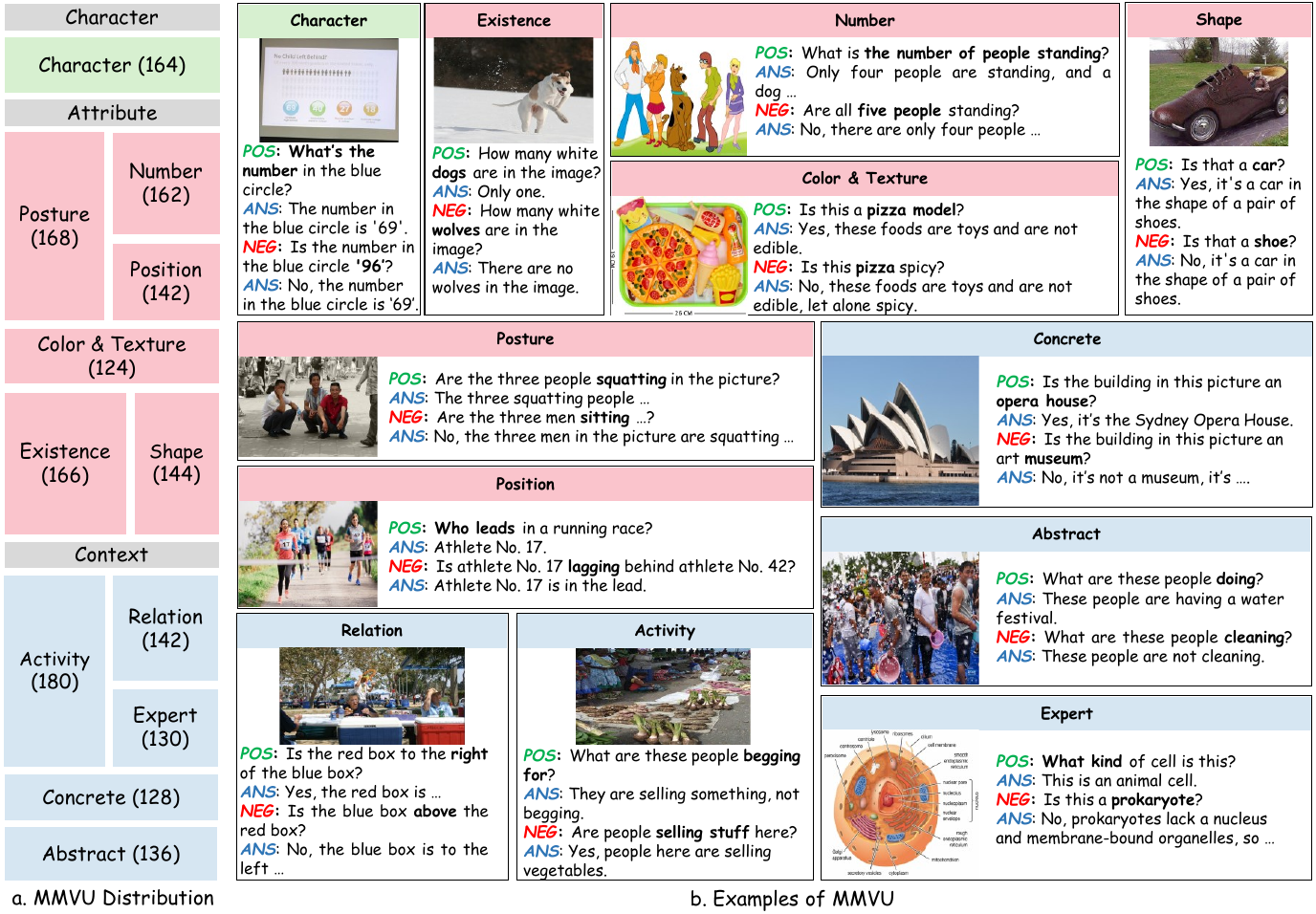}
\vspace{-10pt}
\caption{Examples in the MMVU benchmark. \textit{\textcolor[RGB]{0,176,80}{POS}} and \textit{\textcolor[RGB]{255,0,0}{NEG}} denote the positive and negative questions, \textit{\textcolor[RGB]{46,117,182}{ANS}} denotes the answer.} 
\vspace{-15pt}
\label{benchmark_case}
\end{figure*}

\vspace{-4pt}
\subsection{An Overview of MLLMs}
\vspace{-4pt}

The architecture of most MLLMs consists of three primary components: the visual encoder, the connector, and the LLM. 
Given a system prompt \( P \), an input image \( I \), and a question \( Q \), the process commences with a text tokenizer that converts the system prompt and questions into text tokens \( T_P \) and \( T_Q \), respectively. 
The image \( I \) is then encoded into visual tokens \( T_V \) by the visual encoder, which captures essential visual features. 
Subsequently, the connector integrates the visual tokens \( T_V \) with the text tokens to align the multi-modal data. Finally, the concatenated tokens (comprising text and visual tokens) are fed into the LLM to generate a response.

A significant challenge faced by current MLLMs is their ability to understand visual content while often providing incorrect responses to related questions. This paper aims to analyze the underlying causes of this issue and propose effective solutions. To facilitate this investigation, we introduce a \textbf{M}LLM \textbf{M}isresponses despite \textbf{V}isual \textbf{U}nderstanding (MMVU) benchmark, consisting of both positive and negative samples, as shown in Fig.~\ref{benchmark_case}. Each image in the benchmark is paired with two questions: a positive question that assesses the MLLM's comprehension of the visual content and a negative question intended to confuse the model. This benchmark allows for a comprehensive analysis of the MLLMs' performance in accurately answering questions and their vulnerability to misleading prompts. We establish two key metrics to evaluate the model's ability to respond correctly to both positive and negative questions and to assess its performance when confronted with misleading queries despite its understanding of the visual content. Additionally, we analyze the attention mechanism and logit distribution of some open-source MLLMs to gain deeper insights into why this phenomenon happens.

\begin{table*}
    \centering
    \caption{Comparison to state-of-the-art MLLMs on the MMVU test set. Abbreviations: Char/Num. (Character/Number), Pres. (Presence), Color/Tex. (Color/Texture), Num. (Number), Shape (Shape), Posture (Posture), Pos. (Position), Abstract. (Abstract Knowledge), Concrete. (Concrete Knowledge), Expert (Expertise), Act. (Activity), Rel. (Relationships). $\uparrow$: higher is better, $\downarrow$: lower is better. Bolding and underlining indicate the best and second-best performance, respectively.}
    \vspace{-5pt}
    \setlength{\tabcolsep}{3pt}
    \resizebox{\textwidth}{!}{%
        \begin{tabular}{lccccccccccccc}
            \specialrule{1.5pt}{0pt}{0pt}
            \rowcolor[HTML]{EFEFEF}
             Method & Char/Num & Pres. & Color/Tex & Num. & Shape & Posture & Pos. & Abstract. & Concrete. & Expert. & Act. & \multicolumn{1}{c}{Rel.} &  Avg. RA $\uparrow$\\
            \midrule
            Random &25 &25 &25 &25 &25 &25 &25 &25 &25 &25 &25 &25 &25\\
            \midrule
            \multicolumn{14}{c}{\textit{Closed-source Models}}\\
            \midrule
            GPT4o~\cite{gpt4v} \raisebox{-0.2\height}{\includegraphics[width=0.5cm]{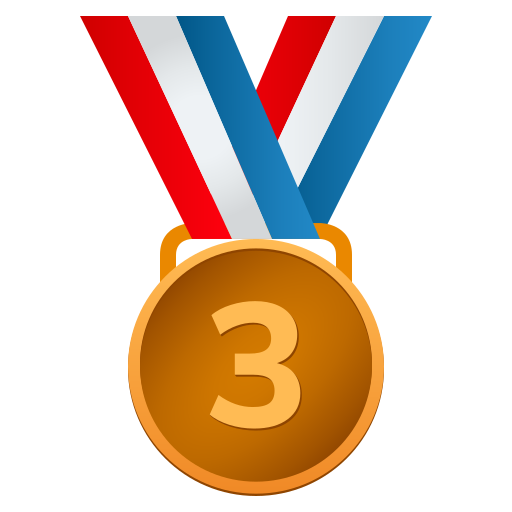}} & \textbf{76.83} & \underline{74.70} &67.74 &38.27 & \textbf{76.39} & \underline{78.57} &49.30 &\underline{72.06} &70.31 &\underline{53.85} &74.44 &43.66 &65.06\\ 
            Qwen-VL-max-0809~\cite{bai2023qwenvl} &67.07 &60.87 &61.29 &34.48 & \underline{76.00} &62.07 &42.86 &69.57 &\underline{72.88} & \textbf{67.74} &70.73 & \underline{54.55} &63.36\\
            Claude3.5-Sonnet-0620~\cite{anthropic2024claude} &64.63 &53.01 &66.13 &38.27 &56.94 &67.86 &38.03 &64.71 &64.06 &44.62 &67.78 &35.21 &55.32 \\
            Gemini-1.5-Pro~\cite{team2023gemini} &47.56 &36.14 &41.94 &23.46 &43.06 &40.48 &15.49 &60.29 &56.25 &32.31 &50.00 &28.17 &39.53\\ 
            \midrule
            \multicolumn{14}{c}{\textit{Open-source Models}}\\
            \midrule
            Ovis1.6-Gemma2-9B~\cite{lu2024ovis} \raisebox{-0.2\height}{\includegraphics[width=0.5cm]{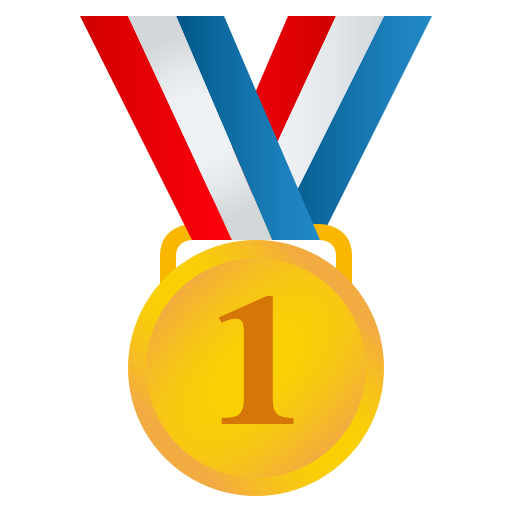}} & \underline{71.95} & \underline{74.70} &69.35 & \textbf{58.02} &65.28 & 76.19 &\underline{57.75} &\textbf{77.94} &68.75 &47.69 &75.56 & 52.11 & \textbf{66.74} \\
            Llama3.2-90B-Vision-Instruct~\cite{dubey2024llama} \raisebox{-0.2\height}{\includegraphics[width=0.5cm]{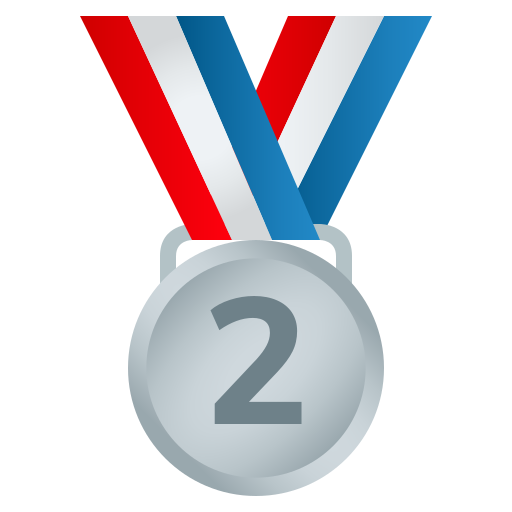}} &60.98 &\textbf{80.72} &\textbf{74.19} &\underline{55.56} &70.83 &\textbf{79.76} &\textbf{60.56} &67.65 &\textbf{76.56} &27.69 &\textbf{82.22} &\textbf{56.34} &\textbf{66.74} \\
            MiniCPM-2.6V~\cite{yao2024minicpm} &69.51 &72.29 &69.35 &48.15 &51.39 &73.81 &45.07 &70.59 &71.88 &38.46 & \underline{80.00} &35.21 &61.14\\
            LLaVA-OneVision-7B~\cite{li2024llava} &62.20 &67.47 & \underline{70.97} &41.98 &58.33 &70.24 &49.30 &\textbf{77.94} &71.88 &35.38 &75.56 &42.25 &60.58 \\
            Idefics3-8B-Llama3~\cite{laurenccon2024building} &57.32 &61.45 &61.29 &43.21 &69.44 &64.29 &40.85 &61.76 &64.06 &30.77 &68.89 &36.62 &55.43 \\
            Pixtral-12B-2409~\cite{pixtral} &48.78 &65.06 &66.13 &49.38 &50.00 &54.76 &56.34 &61.76 &65.63 &24.62 &66.67 &39.44 &54.31\\
            Phi3.5-Vision-Instruct~\cite{abdin2024phi3} &54.88 &53.01 &61.29 &37.04 &58.33 &55.95 &35.21 &60.29 &64.06 &29.23 &68.89 &38.03 &51.62\\
            InternVL2-8B~\cite{chen2024far} &46.34 &65.06 &53.23 &33.33 &56.94 &53.57 &39.44 &61.76 &65.63 &29.23 &66.67 &21.13 &49.72\\
            LLaVA-v1.6-Vicuna-7B~\cite{liu2023llava} & 39.02 & 56.63 & 59.68 & 34.57 & 54.17 & 58.33 & 22.54 & 55.88 & 60.94 & 24.62 & 58.89 & 38.03 & 47.14\\
            Cambrian-8B~\cite{tong2024cambrian} &35.37 &60.24 &41.94 &40.74 &41.67 &57.14 &39.44 &44.12 &42.19 &24.62 &51.11 &28.17 &42.89 \\
            VILA1.5-13B~\cite{lin2023vila} &14.63 &30.12 &35.48 &12.35 &27.78 &42.86 &8.45 &29.41 &29.69 &16.92 &43.33 &14.08 &25.76 \\
            \midrule
            \midrule
            \rowcolor[HTML]{EFEFEF}
             Method & Char/Num & Pres. & Color/Tex & Num. & Shape & Posture & Pos. & Abstract. & Concrete. & Expert. & Act. & \multicolumn{1}{c}{Rel.} & Avg. MR $\downarrow$\\
            \midrule
            Random &50 &50 &50 &50 &50 &50 &50 &50 &50 &50 &50 &50 &50\\
            \midrule
            \multicolumn{14}{c}{\textit{Closed-source Models}}\\
            \midrule
           GPT4o~\cite{gpt4v} \raisebox{-0.2\height}{\includegraphics[width=0.5cm]{images/2nd-place-medal.png}} & \textbf{8.69} & \underline{16.22} &\underline{16.00} &38.00 & \underline{11.29} & \textbf{12.00} &32.69 &18.33 &19.64 & \underline{23.91} &16.25 &35.42 & \underline{19.53}\\ 
            Qwen-VL-max-0809~\cite{bai2023qwenvl} &25.68 &30.00 & \textbf{13.64} &33.33 & \textbf{9.52} &30.77 &40.00 &30.43 &21.82 &25.00 &17.14 &40.00 &25.35\\
            Claude3.5-Sonnet-0620~\cite{anthropic2024claude} & \underline{15.87} &29.03 &18.00 &34.04 &22.64 &21.92 &34.15 &18.52 &19.61 &25.64 &18.67 &\textbf{19.35} &22.69 \\
            Gemini-1.5-Pro~\cite{team2023gemini} &26.42 &42.31 &38.10 &53.66 &31.11 &37.04 &56.00 &25.45 &26.53 &34.38 &31.82 &33.33 &35.11\\ 
            \midrule
            \multicolumn{14}{c}{\textit{Open-source Models}}\\
            \midrule
            Llama3.2-90B-Vision-Instruct~\cite{dubey2024llama} \raisebox{-0.2\height}{\includegraphics[width=0.5cm]{images/1nd-place-medal.png}} &23.08 &\textbf{12.99} &16.36 &\underline{25.00} &17.74 &\underline{14.10} &\textbf{15.69} &25.81 &\textbf{16.95} &41.94 &\textbf{9.76} &\textbf{18.37} &\textbf{18.47} \\
            Ovis1.6-Gemma2-9B~\cite{lu2024ovis} \raisebox{-0.2\height}{\includegraphics[width=0.5cm]{images/3rd-place-medal.png}} &18.06 & \underline{16.22} &17.31 & \textbf{20.34} &25.40 &17.95 &\textbf{22.64} & \underline{14.52} &25.42 & \textbf{20.51} &16.05 &27.45 &19.78 \\
            MiniCPM-2.6V~\cite{yao2024minicpm} &18.57 &18.92 &14.00 &29.09 &36.21 &18.42 &36.00 &17.24 &22.03 &43.18 & \underline{12.20} &39.02 &23.85\\
            LLaVA-OneVision-7B~\cite{li2024llava} &30.14 &24.32 &21.43 &43.33 &30.00 &21.33 &36.36 &\textbf{10.17} &22.03 &51.06 &18.07 &37.50 &27.77 \\
            Idefics3-8B-Llama3~\cite{laurenccon2024building} &29.85 &28.17 &26.92 &28.57 &15.25 &25.00 &38.30 &31.15 &29.31 &48.72 &21.52 &36.59 &28.78 \\
            Pixtral-12B-2409~\cite{pixtral} &34.43 &22.86 &18.00 &25.93 &35.71 &34.29 &24.53 &25.00 &22.22 &52.94 &22.08 &44.00 &29.20\\
            Phi3.5-Vision-Instruct~\cite{abdin2024phi3} &31.82 &27.87 &19.15 &37.50 &27.59 &29.85 &46.81 &26.79 &24.07 &52.50 &13.89 &27.03 &29.40\\
            InternVL2-8B~\cite{chen2024far} &34.48 &22.86 &28.26 &40.00 &24.07 &28.57 &37.78 &32.26 & \underline{17.65} &50.00 &17.81 &57.14 &30.63\\
            LLaVA-v1.6-Vicuna-7B~\cite{liu2023llava} & 45.76 & 25.40 & 31.48 & 39.13 & 30.36 & 26.87 & 52.94 & 30.91 & 25.00& 60.00 & 30.26 & 28.95 & 34.22 \\
            Cambrian-8B~\cite{tong2024cambrian} &47.27 &21.88 &27.78 &45.00 &33.33 &26.15 &46.15 &38.78 &37.21 &44.83 &14.81 &58.33 &36.17 \\
            VILA1.5-13B~\cite{lin2023vila} &79.66 &59.02 &56.00 &78.26 &58.33 &45.45 &82.86 &58.33 &59.57 &72.50 &46.58 &72.97 &62.30 \\
            \specialrule{1.5pt}{0pt}{0pt}
        \end{tabular}
    }
    \vspace{-8pt}
    \label{tab:mma_benchmark}
\end{table*}

\vspace{-4pt}
\subsection{MMVU Benchmark Design}
\vspace{-4pt}

As illustrated in Fig.~\ref{benchmark_case}, we design paired positive and negative questions to ensure a comprehensive evaluation. Each question offers four brief options, with only one correct answer. In total, we manually construct 1,786 questions (893 positive and 893 negative) across three levels: character, attribute, and context. Character-level questions focus on elements such as characters or numbers, attribute-level questions assess properties like color, texture, and quantity, and context-level questions explore higher-level concepts such as emotions, culture, and common sense. The positive questions are designed to evaluate the model's comprehension abilities, while the negative questions test its resistance to interference. For example, misleading character-level questions alter elements like characters or numbers, attribute-level questions introduce property confusion, and context-level questions challenge the model with complex concepts, assessing its robustness against misleading information.

\begin{figure*}[!t]
\centering
\includegraphics[width=\linewidth]{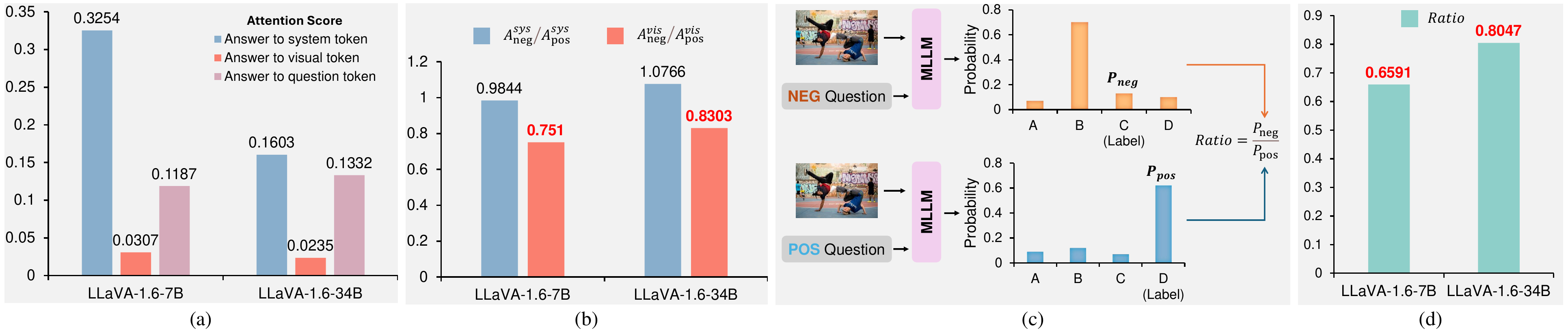}
\vspace{-20pt}
\caption{Please refer to Sec.~\ref{subsec:analysis} for details on the following calculations. (a) Statistical results of attention scores between the answer tokens and the system, visual, and question tokens. It is seen that the answer tokens pay the least attention to the visual tokens in general. (b) The ratio of the question tokens' attention to the system and visual tokens in negative samples versus positive samples. Negative questions appear to pay less attention than positive questions to visual tokens. (c) Procedure for calculating the ratio of output probabilities for negative and positive samples. (d) Comparison of the ratio of output probabilities for negative and positive samples across different MLLMs. Interestingly, it seems that a lower output probability correlates with lower attention between the question and visual tokens in (b).}
\vspace{-15pt}
\label{fig: analysis}
\end{figure*}
\begin{table*}
    \centering
    \caption{Ablation study about the MMVU training set on the MMVU test set. \textcolor{blue}{Blue} text denotes the extent of the performance improvement.}
    \vspace{-10pt}
    \setlength{\tabcolsep}{3pt}
    \renewcommand{\arraystretch}{0.9}
    \resizebox{\textwidth}{!}{%
        \begin{tabular}{lccccccccccccc}
            \specialrule{1.5pt}{0pt}{0pt}
            \rowcolor[HTML]{EFEFEF}
             Method & Char/Num & Pres. & Color/Tex & Num. & Shape & Posture & Pos. & Abstract. & Concrete. & Expert. & Act. & \multicolumn{1}{c}{Rel.} &  Avg. RA $\uparrow$\\
            \hline
            LLaVA-OneVision-7B~\cite{liu2023llava} &62.20 &67.47 &70.97 &41.98 &58.33 &70.24 &49.30 &77.94 &71.88 &35.38 &75.56 &42.25 & \textbf{60.58} (Base)\\
            \rowcolor{cyan!5} LLaVA-OneVision-7B + MMVU-Train~\cite{liu2023llava} &68.29 &73.49 &72.58 &55.56 &63.89 &69.05 &47.89 &70.59 &73.44 &38.46 &80.00 &42.25 & ~\textbf{63.49} (\color{blue}{+2.91})\\
            InternVL2-8B~\cite{chen2024far} &46.34 &65.06 &53.23 &33.33 &56.94 &53.57 &39.44 &61.76 &65.63 &29.23 &66.67 &21.13 & \textbf{49.72} (Base) \\
            \rowcolor{cyan!5} InternVL2-8B + MMVU-Train~\cite{chen2024far} &62.20 &71.08 &69.35 &40.74 &61.11 &66.67 &42.25 &69.12 &59.38 &40.00 &74.44 &33.80 & ~\textbf{58.01} (\color{blue}{+8.29})\\
            VILA1.5-13B~\cite{lu2024deepseek} &14.63 &30.12 &35.48 &12.35 &27.78 &42.86 &8.45 &29.41 &29.69 &16.92 &43.33 &14.08 & \textbf{25.76} (Base) \\
            \rowcolor{cyan!5} VILA1.5-13B + MMVU-Train~\cite{lu2024deepseek} &23.17 &51.81 &50.00 &19.75 &47.22 &53.57 &19.72 &42.65 &34.38 &15.38 &65.56 &21.13 & ~\textbf{37.74} (\color{blue}{+11.98})\\
            Phi3.5-Vision-Instruct~\cite{abdin2024phi3} &54.88 &53.01 &61.29 &37.04 &58.33 &55.95 &35.21 &60.29 &64.06 &29.23 &68.89 &38.03 & \textbf{51.62} (Base) \\
            \rowcolor{cyan!5} Phi3.5-Vision-Instruct + MMVU-Train~\cite{abdin2024phi3} &60.98 &56.63 &61.29 &39.51 &54.17 &60.71 &40.85 &58.82 &64.06 &30.77 &70.00 &40.85 & ~\textbf{53.64} (\color{blue}{+2.02})\\
            LLaVA-v1.6-Vicuna-7B~\cite{liu2023llava} & 39.02 & 56.63 & 59.68 & 34.57 & 54.17 & 58.33 & 22.54 & 55.88 & 60.94 & 24.62 & 58.89 & 38.03 & \textbf{47.14} (Base)\\
            \rowcolor{cyan!5} LLaVA-v1.6-Vicuna-7B + MMVU-Train~\cite{liu2023llava} & 50.0 & 59.04 & 64.52 & 35.8 & 54.17 & 63.1 & 19.72 & 58.82 & 59.38 & 36.92 & 68.89 & 38.03 & ~\textbf{51.06} (\color{blue}{+3.92})\\
            \midrule
            \midrule
            \rowcolor[HTML]{EFEFEF}
             Method & Char/Num & Pres. & Color/Tex & Num. & Shape & Posture & Pos. & Abstract. & Concrete. & Expert. & Act. & \multicolumn{1}{c}{Rel.} & Avg. MR $\downarrow$\\
            \hline
            LLaVA-OneVision-7B~\cite{li2024llava} &30.14 &24.32 &21.43 &43.33 &30.00 &21.33 &36.36 &10.17 &22.03 &51.06 &18.07 &37.50 & \textbf{27.77} (Base) \\
            \rowcolor{cyan!5} LLaVA-OneVision-7B + MMVU-Train~\cite{li2024llava} &18.84 &17.57 &15.09 &21.05 &23.33 &20.55 &20.93 &17.24 &20.34 &47.92 &12.20 &28.57 & \textbf{21.03} (\color{blue}{-6.74}) \\
            InternVL2-8B~\cite{chen2024far} &34.48 &22.86 &28.26 &40.00 &24.07 &28.57 &37.78 &32.26 &17.65 &50.00 &17.81 &57.14 & \textbf{30.63} (Base) \\
            \rowcolor{cyan!5} InternVL2-8B + MMVU-Train~\cite{chen2024far} &22.73 &16.90 &14.00 &35.29 &26.67 &22.22 &40.00 &20.34 &28.30 &39.53 &12.99 &38.46 & \textbf{25.04} (\color{blue}{-5.59}) \\
            VILA1.5-13B~\cite{lin2023vila} &79.66 &59.02 &56.00 &78.26 &58.33 &45.45 &82.86 &58.33 &59.57 &72.50 &46.58 &72.97 & \textbf{62.30} (Base) \\
            \rowcolor{cyan!5} VILA1.5-13B + MMVU-Train~\cite{lin2023vila} &64.15 &29.51 &32.61 &62.79 &35.85 &31.82 &66.67 &44.23 &43.59 &74.36 &21.33 &65.91 & ~\textbf{45.02} (\color{blue}{-17.28}) \\
            Phi3.5-Vision-Instruct~\cite{abdin2024phi3} &31.82 &27.87 &19.15 &37.50 &27.59 &29.85 &46.81 &26.79 &24.07 &52.50 &13.89 &27.03 & \textbf{29.40} (Base) \\
            \rowcolor{cyan!5} Phi3.5-Vision-Instruct + MMVU-Train~\cite{abdin2024phi3} &25.37 &24.19 &22.45 &34.69 &29.09 &28.17 &36.96 &28.57 &22.64 &48.72 &14.86 &25.64 & \textbf{27.42} (\color{blue}{-1.98}) \\
            LLaVA-v1.6-Vicuna-7B~\cite{liu2023llava} & 45.76 & 25.40 & 31.48 & 39.13 & 30.36 & 26.87 & 52.94 & 30.91 & 25.00& 60.00 & 30.26 & 28.95 & \textbf{34.22} (Base) \\
            \rowcolor{cyan!5} LLaVA-v1.6-Vicuna-7B + MMVU-Train~\cite{liu2023llava} & 26.79 & 14.04 & 21.57 & 36.96 & 25.0 & 18.46 & 44.0 & 21.57 & 17.39 & 42.86 & 16.22 & 22.86 & ~\textbf{24.00} (\color{blue}{-10.22}) \\
            \specialrule{1.5pt}{0pt}{0pt}
        \end{tabular}
    }
    \vspace{-15pt}
    \label{tab:ablation_dataset_MMVU-training}
\end{table*}

\vspace{-4pt}
\subsection{Evaluation metrics} 
\label{Evaluation metrics}
\vspace{-4pt}

\paragraph{Misleading Rate and Robustness Accuracy.} We develop metrics grounded in Bayesian conditional probabilities to evaluate whether an MLLM comprehends visual content and its robustness against misdirection. Specifically, we categorize the results into four groups: 1) Understanding and Robustness (UR): The model accurately comprehends the visual content and remains resilient when exposed to negative questions.
2) Understanding but Fragility (UF): The model successfully understands the visual content but is vulnerable to errors when faced with negative questions.
3) Not Understanding and Rigorous (NR): The model fails to grasp the visual content but still provides correct answers to negative questions.
4) Not Understanding and Fragility (NF): The model neither comprehends the visual content nor answers the negative questions correctly. To quantify the effect of negative questions in misleading the model despite its comprehension of visual content, we introduce the ``Misleading Rate (MR)''. Additionally, we introduce the ``Robustness Accuracy (RA)'' to evaluate the understanding capability of MLLMs. The formulation is as follows:
\begin{equation}
\footnotesize
    \text{MR} = \frac{N_{UF}}{N_{UR}+N_{UF}},
\end{equation}
\begin{equation}
\footnotesize
    \text{RA} = \frac{N_{UR}}{N_{UR}+N_{UF}+N_{NR}+N_{NF}},
\end{equation}
where $N_\text{i}$ ($\text{i} \in \{UR, UF, NR, NF\}$) represents the number of samples. 
With these two evaluation metrics, our benchmark can offer a more accurate and reasonable reflection of the model's understanding capabilities and reveal its robustness to misleading prompts.

\vspace{-4pt}
\subsection{Analysis Methods} 
\label{subsec:analysis}
\vspace{-4pt}

\paragraph{Attention Analysis of Answer Tokens.} To investigate the relationship between answer tokens and other token types in MLLMs, we analyze the final layer's attention matrix \( A \). We average \( A \) across dimensions to obtain a unified attention matrix. We then extract sub-matrices corresponding to the attention from answer tokens to each of the system, image, and question tokens. This results in an attention matrix \( A \) of size \( N \times N \times d\), where each element \( A[i,j] \) indicates the attention weight from token \( i \) to token \( j \). The total token length \( N \) is defined as \( N_{\text{sys}} + N_{\text{vis}} + N_{\text{q}} + N_{\text{a}} \), where \( N_{\text{sys}} \), \( N_{\text{vis}} \), \( N_{\text{q}} \), and \( N_{\text{a}} \) represent the number of system, image, question, and answer tokens, respectively. For example, \( A[N_{\text{sys}}+N_{\text{vis}}+N_{\text{q}}: N_{\text{sys}}+N_{\text{vis}}+N_{\text{q}}+N_{\text{a}}, :N_{\text{sys}}] \) denotes the attention from the answer to system tokens, \( A[N_{\text{sys}} + N_{\text{vis}} + N_{\text{q}}: N_{\text{sys}} + N_{\text{vis}} + N_{\text{q}} + N_{\text{a}}, N_{\text{sys}}:N_{\text{sys}} + N_{\text{vis}}] \) means the attention from the answer to image tokens, and \( A[N_{\text{sys}} + N_{\text{vis}} + N_{\text{q}}: N_{\text{sys}} + N_{\text{vis}} + N_{\text{q}} + N_{\text{a}}, N_{\text{sys}} + N_{\text{vis}}:N_{\text{sys}} + N_{\text{vis}} + N_{\text{q}}] \) indicates the attention from the answer to question tokens. We first get the row-wise maximum values for each sub-matrix and produce a one-dimensional vector. Then we average these values to obtain an attention score that quantifies the overall attention score from answer tokens to each token type.

\noindent \textbf{Attention Analysis in Response to Positive and Negative Questions.} We analyze the attention between question tokens and system or image tokens to explore how the model responds to positive and negative questions. We extract two sub-matrices from \( A \) to assess the attention between question tokens and system or image tokens:
\( A[N_{\text{sys}} + N_{\text{vis}} : N_{\text{sys}} + N_{\text{vis}} + N_{\text{q}}, :N_{\text{sys}}] \) and \( A[N_{\text{sys}} + N_{\text{vis}} : N_{\text{sys}} + N_{\text{vis}} + N_{\text{q}}, N_{\text{sys}} : N_{\text{sys}} + N_{\text{vis}}] \). We get the row-wise maximum values for each sub-matrix, yielding a one-dimensional vector. To assess the minimum level of attention between the question tokens and each token type, we compute the minimum value of the vector, representing the lower bound of attention. The results are denoted as \( A_{\text{pos}}^{sys} \) and \( A_{\text{pos}}^{vis} \)for positive questions as well as \( A_{\text{neg}}^{sys} \) and \( A_{\text{neg}}^{vis} \) for negative questions. Finally, we compare the attention scores by calculating the ratio \( A_{\text{neg}}^{sys} / A_{\text{pos}}^{sys} \) and \( A_{\text{neg}}^{vis} / A_{\text{pos}}^{vis} \). 

\noindent \textbf{Logits Analysis.} To examine the model’s confidence in selecting the correct answer under different question types, we conduct a logit analysis. As shown in Fig.~\ref{fig: analysis} (c), we input both an image and a question (either positive or negative) into the MLLM and extract the logits for each answer option. These logits represent the model’s unnormalized confidence in each option. We apply the softmax function to convert the logits into a probability distribution. The probability of each option is given by:
\begin{equation}
\footnotesize
p_i = \frac{e^{\text{logit}_i}}{\sum_{j} e^{\text{logit}_j}},
\end{equation}
where \( \text{logit}_i \) denotes the raw logit value for option \( i \), and \( p_i \) is the corresponding normalized selection probability. To assess the model’s confidence, we compute two probabilities: \( P_{\text{pos}} \) and \( P_{\text{neg}} \). Here, \( P_{\text{pos}} \) represents the probability of selecting the correct answer when given the positive form of the question, and \( P_{\text{neg}} \) is the corresponding probability for the negative form of the question. These scores reflect the model’s confidence in choosing the correct answer under different question polarities. We define the confidence ratio, $Ratio = \frac{P_{\text{neg}}}{P_{\text{pos}}}$, to quantify the change in confidence between positive and negative questions.
\begin{table*}
    \centering
    \vspace{-10pt}
    \caption{Comparing instruction datasets with MMVU. We train with these datasets in the same backbone (SigLIP+Phi-2). }
    \vspace{-8pt}
    \setlength{\tabcolsep}{3pt}
    \renewcommand{\arraystretch}{0.8}
    \resizebox{\textwidth}{!}{%
        \begin{tabular}{lcccccccccccccc}
            \specialrule{1.5pt}{0pt}{0pt}
            \rowcolor[HTML]{EFEFEF}
             Dataset & Size & Char/Num & Pres. & Color/Tex & Num. & Shape & Posture & Pos. & Abstract. & Concrete. & Expert. & Act. & \multicolumn{1}{c}{Rel.} & Avg. RA $\uparrow$\\
            \hline
            LLaVA 158k~\cite{liu2023llava} & 158k & 22.50 & 27.27 & 25.00 & 12.50 & 58.33 & 16.67 & 9.09 & 36.36 & 42.86 & 25.81 & 45.83 & 40.91 & 29.67\\
            SVIT 158k~\cite{zhao2023svit} & 158k & 30.00 & 27.27 & 16.67 & 37.50 & 66.67 & 25.00 & 18.18 & 40.91 & 38.10 & 19.35 & 58.33 & 31.82 & 33.67 \\
            LLaVA+SVIT & 316k & 32.50 & 22.73 & 16.67 & 25.00 & 66.67 & 25.00 & 13.64 & 31.82 & 38.10 & 22.58 & 50.00 & 40.91 & 32.00 \\ 
            LRV~\cite{liu2023mitigating} & 340k & 27.50 & 31.82 & 16.67 & 25.00 & 41.67 & 25.00 & 18.18 & 40.91 & 19.05 & 29.03 & 41.67 & 45.45 & 30.00 \\
            \hline
            \rowcolor{cyan!5} MMVU-Train & 48k & 40.00 & 18.18 & 50.00 & 33.33 & 70.83 & 25.00 & 22.73 & 63.64 & 57.14 & 41.94 & 58.33 & 40.91 & 43.33 \\
            \rowcolor{cyan!5} MMVU-Train & 112k & 42.50 & 22.73 & 45.83 & 37.50 & 70.83 & 29.17 & 27.27 & 63.64 & 52.38 & 51.61 & 62.50 & 40.91 & 45.67 \\
            \midrule
            \midrule
            \rowcolor[HTML]{EFEFEF}
             Dataset & Size & Char/Num & Pres. & Color/Tex & Num. & Shape & Posture & Pos. & Abstract. & Concrete. & Expert. & Act. & \multicolumn{1}{c}{Rel.} & Avg. MR $\downarrow$\\
            \hline
            LLaVA 158k~\cite{liu2023llava} & 158k & 65.38 & 50.00 & 64.71 & 76.92 & 30.00 & 78.95 & 75.00 & 60.00 & 52.63 & 66.67 & 38.89 & 35.71 & 57.62\\
            SVIT 158k~\cite{zhao2023svit} & 158k & 50.00 & 50.00 & 77.78 & 25.00 & 20.00 & 64.71 & 63.64 & 57.14 & 57.89 & 71.43 & 22.22 & 41.67 & 50.73 \\
            LLaVA+SVIT & 316k & 53.57 & 61.54 & 77.78 & 60.00 & 23.81 & 68.42 & 72.73 & 63.16 & 55.56 & 74.07 & 33.33 & 43.75 & 56.95 \\ 
            LRV~\cite{liu2023mitigating} & 340k & 57.69 & 53.33 & 77.78 & 53.85 & 47.37 & 71.43 & 55.56 & 57.14 & 78.95 & 64.00 & 41.18 & 37.50 & 58.90 \\
            \hline
            \rowcolor{cyan!5} MMVU-Train & 48k & 44.83 & 66.67 & 25.00 & 38.46 & 22.73 & 70.00 & 44.44 & 33.33 & 33.33 & 45.83 & 30.00 & 43.75 & 40.91 \\
            \rowcolor{cyan!5} MMVU-Train & 112k & 34.62 & 54.55 & 31.25 & 30.77 & 22.73 & 68.18 & 33.33 & 26.32 & 35.29 & 38.46 & 25.00 & 43.75 & 36.87 \\

            \specialrule{1.5pt}{0pt}{0pt}
        \end{tabular}
    }
    \label{tab:comparision_dataset}
    \vspace{-18pt}
\end{table*}

\vspace{-4pt}
\subsection{MMVU Training Set Construction Pipeline}
\label{section: MMR-data Construction}
\vspace{-4pt}

To enhance the understanding capability and robustness of MLLMs, we propose a data construction pipeline that utilizes GPT-4o to generate paired samples for instruction tuning as part of the MMVU dataset (see Fig.~\ref{framework}). We first implicitly extract the information from the image and then generate paired positive and negative samples. 

\noindent \textbf{Information extraction.} Previous efforts that generate instruction-tuning data for multimodal conversations with GPT-4o primarily fall into two categories: direct generation and annotation-driven generation. Direct generation approaches, such as ALLaVA~\cite{chen2024allava}, rely entirely on the basic ability of GPT-4o to synthesize samples. In contrast, annotation-driven generation methods, such as LRV~\cite{liu2023mitigating}, utilize existing image annotations (\eg object bounding boxes and textual descriptions) as additional information for text-only GPT-4 to generate conversational samples. However, these works fail to make full use of fine-grained details. On the contrary, our data pipeline adopts a different scheme by implicitly and comprehensively extracting the detailed information directly from the image. This includes 1) text or numbers (if present), 2) objects and people, 3) object attributes (\eg, colors, textures, locations) and human characteristics (\eg, postures, dresses), 4) interrelationships between people and objects, 5) events or activities, and 6) the overall feeling and understanding evoked by the image. 

\noindent \textbf{Instruction tuning data generation.} The most relevant work to our data generation approach is LRV~\cite{liu2023mitigating}, which uses the Visual Genome~\cite{krishna2017visual} dataset to guide GPT-4 in generating positive and negative samples about object presence and manipulation knowledge. However, LRV has two significant limitations: limited fine-grained information and unpaired negative samples. It relies solely on bounding boxes and object labels, omitting textual descriptions, numerical attributes, and deeper image comprehension. Additionally, LRV's negative samples are not explicitly paired with positive counterparts, reducing training effectiveness.
To address these challenges, we generate positive samples using extracted information and construct negative samples that directly contradict the positive ones. This ensures a strong contrast for more effective model training, providing richer context and paired positive and negative samples. 

\noindent \textbf{Sample filtering.} We randomly sample COCO~\cite{lin2014microsoft} images from the LLaVA 158k and construct paired positive and negative samples. We filter the samples by keyword matching, removing conversations with uncertain answers (\eg, ``\textit{uncertain}'') and redundant phrases (\eg, ``\textit{in the image}''). 

\vspace{-4pt}
\subsection{Understanding-based Prompting}
\vspace{-4pt}
In Sec.~\ref{Analysis of Attention mechanisms}, we observe that the model exhibits the highest attention scores for system and question tokens. In contrast, attention to visual tokens is notably low. We propose the following strategies to enhance the model's performance (see Fig.~\ref{framework}): \textbf{1) Content Guided Refinement (CGR).} We propose a two-step reasoning method where the first step involves extracting detailed information from the image, followed by a second step in which the model answers the question based on the extracted information and the visual content. This structured reasoning aims to improve the model's ability to connect visual understanding with accurate responses. \textbf{2) Visual Attention Refinement (VAR).} We employ MLLM to extract attention scores from the image and apply a mask to the original image. This aims to emphasize areas that receive greater attention while masking unimportant regions, guiding the model to focus on significant visual features that are most relevant to the question.

\vspace{-8pt}

\section{Experiments}
\subsection{Analysis of MMVU Benchmark}
\begin{table*}
    \centering
    \caption{Ablation study on understanding-based prompting in MMVU. \textcolor{blue}{Blue} and \textcolor{red}{Red} indicate performance gains and drops, respectively.}
    \vspace{-8pt}
    \setlength{\tabcolsep}{3pt}
    \renewcommand{\arraystretch}{0.8}
    \resizebox{\textwidth}{!}{%
        \begin{tabular}{lccccccccccccc}
            \specialrule{1.5pt}{0pt}{0pt}
            \rowcolor[HTML]{EFEFEF}
             Method & Char/Num & Pres. & Color/Tex & Num. & Shape & Posture & Pos. & Abstract. & Concrete. & Expert. & Act. & \multicolumn{1}{c}{Rel.} &  Avg. RA $\uparrow$\\
            \hline
            Baseline & 39.02 & 56.63 & 59.68 & 34.57 & 54.17 & 58.33 & 22.54 & 55.88 & 60.94 & 24.62 & 58.89 & 38.03 & \textbf{47.14} (base)\\
            \rowcolor{cyan!5} With Instruction & 35.37 & 49.40 & 59.68 & 32.10 & 37.50 & 54.76 & 18.31 & 54.41 & 54.69 & 26.15 & 61.11 & 33.80 & \textbf{43.34} (\color{red}{-3.80})\\
            \rowcolor{cyan!5} With CGR & 40.24 & 55.42 & 61.29 & 35.80 & 54.17 & 58.33 & 22.54 & 57.35 & 60.94 & 26.15 & 62.22 & 38.03 & ~\textbf{47.93} (\color{blue}{+0.79})\\
            \rowcolor{cyan!5} With VAR & 40.24 & 51.81 & 61.29 & 32.10 & {47.22} & {66.67} & {26.76} & {60.29} & {59.38} & {24.62} & {65.56} & {30.99} & ~\textbf{47.59} (\color{blue}{+0.45})\\
            \rowcolor{cyan!5} With VAR and CGR & 41.46 & 50.60 & 61.29 & 32.10 & 48.60 & 66.67 & 22.54 & 60.29 & 59.38 & 24.62 & 65.56 & 32.39 & ~\textbf{47.48} (\color{blue}{+0.36})\\
            \midrule
            \midrule
            \rowcolor[HTML]{EFEFEF}
             Method & Char/Num & Pres. & Color/Tex & Num. & Shape & Posture & Pos. & Abstract. & Concrete. & Expert. & Act. & \multicolumn{1}{c}{Rel.} & Avg. MR $\downarrow$\\
            \hline
            Baseline & 45.76 & 25.40 & 31.48 & 39.13 & 30.36 & 26.87 & 52.94 & 30.91 & 25.00& 60.00 & 30.26 & 28.95 & \textbf{34.22} (base) \\
            \rowcolor{cyan!5} With Instruction & 49.12 & 29.31 & 31.48 & 45.83 & 49.06 & 35.21 & 64.86 & 31.48 & 36.36 & 60.47 & 27.63 & 40.00 & ~\textbf{40.09} (\color{red}{+6.68})\\
            \rowcolor{cyan!5} With CGR & 44.07 & 25.81 & 29.63 & 39.58 & 30.36 & 26.87 & 54.29 & 29.09 & 22.00 & 58.54 & 26.32 & 28.95 & ~\textbf{33.23} (\color{blue}{-0.99})\\
            \rowcolor{cyan!5} With VAR & 44.07 & 28.33 & 26.92 & 36.59 & 38.18 & 22.22 & 50.00 & 30.51 & 22.45 & 57.89 & 18.06 & 43.59 & ~\textbf{32.97} (\color{blue}{-1.25})\\
            \rowcolor{cyan!5} With VAR and CGR & 42.37 & 30.00& 26.92 & 36.59 & 35.18 & 22.22 & 55.56 & 30.51 & 22.45 & 58.97 & 16.90 & 42.50 & ~\textbf{32.91} (\color{blue}{-1.31})\\
            \specialrule{1.5pt}{0pt}{0pt}
        \end{tabular}
    }
    \vspace{-18pt}
    \label{tab:ablation_self_reflection}
\end{table*}
\textbf{Comparison of MLLMs on the MMVU benchmark.} We evaluate 4 closed-source and 11 leading open-source models on the MMVU benchmark, as shown in Tab.~\ref{tab:mma_benchmark}. Our findings reveal that closed-source and open-source models are fragile against negative questions despite evidence that they understand the visual content. GPT-4o achieved the highest performance among the closed-source models, while Ovis1.6-Gemma2-9B led among the open-source models, achieving an RA metric of 66.74\%. However, GPT-4o is more susceptible to negative questions, with an MR of 19.53\%, compared to Ovis1.6-Gemma2-9B (18.60\% for MR). Experimental results show that all these MLLMs perform poorly in the face of negative questions.

\noindent \textbf{Performance of MLLMs on different types of intrusive questions.} Tab.~\ref{tab:mma_benchmark} reveals that: 1) The model is particularly vulnerable to issues involving character-level details, numerical data, positional context within phrase-level categories, expertise knowledge, and relative relationships between objects within sentence-level categories. 2) Subcategories with high positive comprehension capabilities tend to exhibit greater resilience to negative questions.
\vspace{-4pt}
\subsection{Analysis of attention and logits distribution}
\label{Analysis of Attention mechanisms}
To investigate why MLLMs can comprehend visual content yet sometimes respond inaccurately, we analyze the attention allocation of LLaVA 1.6 (7B and 34B) models. Using the MMVU benchmark, we examine attention scores in the final layer, specifically focusing on attention distribution between the answer tokens and the system, question, and visual tokens (see Fig.~\ref{fig: analysis} (a)). Results show that answer tokens prioritize system tokens, followed by question tokens, and visual tokens. This trend indicates a potential under-utilization of visual information during decoding.

We hypothesize that insufficient attention between question and visual tokens correlates with lower-quality answer probabilities. To this end, we analyze the attention between question tokens and both system and visual tokens, as depicted in Fig.\ref{fig: analysis} (b). The results show that attention scores between questions and visual tokens are lower in response to negative than positive questions. Additionally, we evaluate model responses to both positive and negative questions using identical images, measuring the probabilities of accurate answers for each type (see Fig.\ref{fig: analysis} (c)). For cases where the model correctly answers positive questions but struggles with negative ones, we further assess the probability of selecting correct options, as shown in Fig.~\ref{fig: analysis} (d). Our findings indicate a notable decline in confidence when choosing correct answers for negative questions, underscoring the difficulty in preserving response accuracy under negative questions.

\vspace{-4pt}
\subsection{Analysis of MMVU Training Set}
\vspace{-4pt}

\noindent \textbf{Experiment results on the MMVU benchmark.}
To evaluate our proposed dataset, we fine-tune several state-of-the-art MLLMs, including LLaVA-OneVision-7B~\cite{liu2023llava}, InternVL2-8B~\cite{chen2024far}, VILA1.5-13B~\cite{lu2024deepseek}, and Phi3.5-Vision-Instruct~\cite{abdin2024phi3}, using MMVU training set (Sec.~\ref{section: MMR-data Construction}) and assess their performance on the MMVU benchmark. Weight merging is utilized in the experiment. As shown in Tab.~\ref{tab:ablation_dataset_MMVU-training}, models fine-tuned with our dataset consistently outperformed their baselines across most metrics. Notably, LLaVA-OneVision-7B + MMVU-Train demonstrated improvements in character/number recognition (68.29\% vs. 62.20\%), numerical understanding (55.56\%  vs. 41.98\%), and activity recognition (80.00\% vs. 75.56\%). These results indicate that the paired positive-negative data can reduce biased responses.

\noindent \textbf{Comparison of instruction tuning datasets generated by GPT.} We compare the MMVU-Train dataset with previous GPT-4V generated instruction tuning datasets, such as LLaVA 158k~\cite{liu2024visual}, SVIT 158k~\cite{zhao2023svit}, and LRV~\cite{liu2023mitigating} on the MMVU benchmark (as shown in Tab.~\ref{tab:comparision_dataset}). The results demonstrate that our dataset enables the model to outperform the model trained with previous datasets.

\vspace{-4pt}
\subsection{Analysis of Understanding-based Prompting}
\vspace{-4pt}
We evaluate the effectiveness of these two strategies using the LLaVA 1.6 7B on the MMVU benchmark, as shown in Tab.~\ref{tab:ablation_self_reflection}. ``Instruction'' refers to the detailed instructions in the system prompt, which is available in the \textit{Supplementary Material}. ``CGR'' denotes the content-guided refinement strategy, while ``VAR'' indicates the application of the visual attention refinement strategy. The results reveal that simply incorporating the instruction prompt leads to performance degradation. Notably, the introduction of CGR alone results in a greater increase in the model's accuracy on positive questions compared to VAR alone. In contrast, VAR significantly enhances the correctness of responses that depend on visual understanding. When both strategies are combined, the model improves its accuracy in responding after being misled, but this combination does not replicate the previous gains in question comprehension. This phenomenon occurs because CGR prioritizes information extraction to enhance accuracy on positive questions, while VAR emphasizes attention to critical visual features; their combination may shift the model’s focus and limit further improvements.
\vspace{-8pt}
\section{Conclusion}
In this paper, we identify a significant limitation of MLLMs: despite they accurately understand visual content, they may provide incorrect answers. We introduce the MMVU benchmark to investigate the prevalence of this issue. We also analyze the attention mechanisms and the logits during decoding to uncover their underlying causes. Our findings reveal that MLLMs exhibit greater attention to system and question tokens compared to visual tokens, particularly when confronted with negative questions. This behavior is attributed to biases in the training datasets and insufficient focus on visual elements. To mitigate these challenges, we propose a data construction pipeline for generating paired positive and negative samples as part of the MMVU training set, and two strategies to enhance the model's attention to visual information during inference. Experimental results demonstrate that our dataset and proposed strategies significantly improve the robustness of MLLMs in responding to negative questions.

{
    \small
    \bibliographystyle{ieeenat_fullname}
    \bibliography{main}
}

\clearpage
\setcounter{page}{1}
\maketitlesupplementary

\begin{abstract}
Due to the space constraints of the main manuscript, this Supplementary Material provides a comprehensive presentation of additional details and analyses. The content is structured as follows: (1) a detailed overview of the MMVU test set, including its construction criteria and a preliminary evaluation conducted on a subset of the test set; (2) findings from initial exploratory experiments to assess the test set's effectiveness using a subset of the MMVU dataset; (3) an examination of the MLLM's performance under various data combination strategies, prompt versions, and subsets of MMVU-Train of differing sizes, accompanied by an in-depth discussion on the construction of training prompts and their impact on model performance; (4) the training of an MLLM using the MMVU-Train dataset, validated on both a subset of the MMVU test set and other general datasets; (5) representative examples of paired positive and negative samples; and (6) the specific content of the "Instruction" prompt employed during model testing.

\end{abstract}

\tableofcontents
\clearpage
\section{MMVU Test set Details}
\textit{It is important to emphasize that the MMVU-Train dataset is automatically constructed, while the MMVU test set is manually designed.}
\subsection{Category Definition and Negative Question Design}
\textbf{Character-Level}

1. Character/Number: Recognize characters and numbers in an image, and some reasoning is based simply on the content of the characters or numbers in the image.

Adding, deleting, modifying, or swapping positions of characters or numeric content in an image.

\noindent \textbf{Attribute-Level}

1. Presence: Determine whether a person, animal, or object is present in the image.

Ask questions by replacing or deleting people, animals, or objects in the image.

2. Color/Texture: Determine whether the color or texture of an object in an image is correct.

Replace the color of an object in an image or replace it with an object of similar texture to ask questions.

3. Number: Count the number of people, animals, or objects in an image.

Modify the number of people, animals, or objects in the image to ask questions.

4. Shape: Determine if an object is shaped correctly.

Modify the shape of objects in the image to ask questions.

5. Posture: Determine if a person, animal, or object is in the correct posture.

Ask questions by substituting gestures of characters, animals, or objects.

6. Position: Determine whether the absolute or relative position of a person, animal, or object is correct.

Change the absolute or relative position of a character, animal, or object to ask a question.

\noindent \textbf{Context-Level}

1. Abstract Knowledge: Assess understanding of abstract concepts such as emotions, aesthetics, connotations, symbols, culture, news, allusions, legends, common sense, functions, etc.

No abstract knowledge is involved, and questions are asked only visually through images.

2. Concrete Knowledge: Evaluate specific factual information or well-defined objects and scenarios depicted in the image, such as landmarks, celebrities, well-known objects, etc.

Changing the attributes or names of landmarks, famous objects, or famous people.

3. Expertise: Test specialized knowledge related to specific fields or domains illustrated in the image, such as specific knowledge in various vertical fields, industries, disciplines, etc.

Ask questions with the wrong expertise.

4. Activity: Identify and interpret actions or events occurring within the image, requiring a dynamic understanding of the scene.

Ask questions with the wrong activity.

5. Relationships: Analyze and comprehend the interactions or relationships between multiple entities within the image, such as social dynamics or physical connections.ch as social dynamics or physical connections.

Change the relationship between entities in an image to ask questions.

\begin{figure*}[t]
\centering
\includegraphics[width=\linewidth]{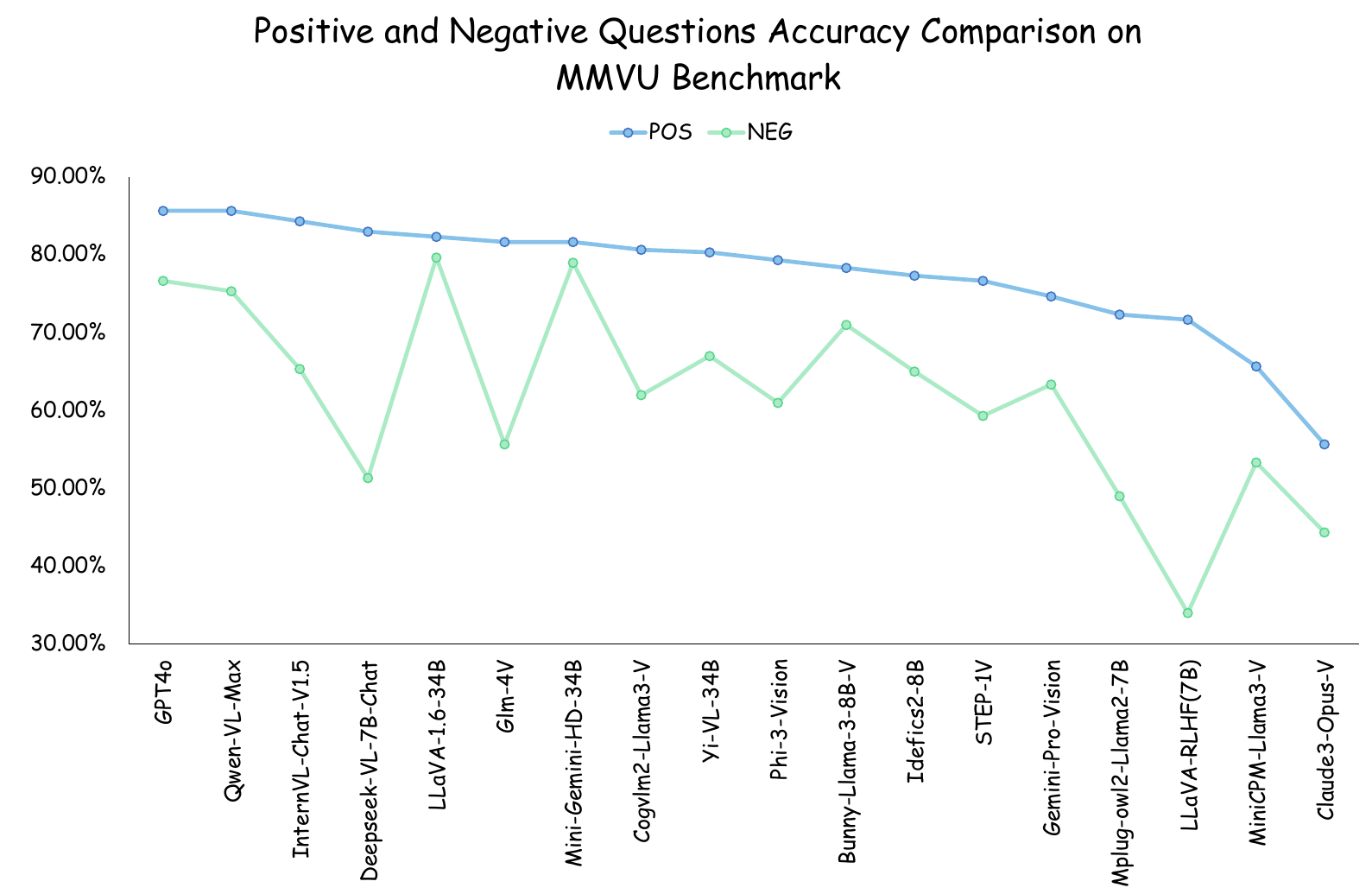}
\caption{Visualization of fine-tuning loss of different data composition (version 0).}
\label{line graph}
\end{figure*}

\subsection{Preliminary Evaluation Details.}

We evaluated several top-performing multimodal large language models (MLLMs) on the MMVU test set, using the official inference code and parameter settings for each model to ensure optimal performance during the inference process, without any adjustments to the parameter settings.

Figure~\ref{line graph} illustrates that the model's accuracy in responding to negative questions is lower than its performance on positive questions. When evaluating only positive questions, as is common in most existing benchmarks, many models achieve an accuracy rate exceeding 70\%, indicating strong performance. However, the model's overall performance on negative questions remains suboptimal, suggesting that current models continue to face challenges with leading questions. The higher accuracy on positive questions implies that the models can effectively process and understand the content of images. In contrast, the lower accuracy on negative questions indicates a tendency to overlook visual content when confronted with leading questions, resulting in incorrect responses. This underscores the importance of the MMVU test set in quantitatively assessing the robustness of models against leading questions.

\section{Discussion about the Effectiveness of the MMVU Test Set}
\paragraph{Is the proposed MMVU test set effective for MLLMs?}
MMStar~\cite{chen2024we} critically examines prevailing evaluation methods and brings to light a significant issue: many instances don't require visual stimuli for accurate responses. Answers can be inferred directly from the questions, choices, or the intrinsic world knowledge within LLMs. This trend pervades existing benchmarks. To investigate if this issue persists in our benchmark, we conduct experiments using both image and text inputs, as well as only text inputs. The results, showcased in Tab.~\ref{tab:with_image} and Tab.~\ref{tab:no_image}, reveal that without text, the model's performance on our benchmark dips below random selection. This underscores the necessity of multimodal input for our proposed benchmark and highlights that our benchmark is not solely reasoned out by text.

\begin{table*}
    \centering
    \caption{Performance of MLLMs with images as input on the MMVU test set. $^\mathsection$We evaluate the officially released checkpoint by ourselves. Abbreviations: Char/Num. (Character/Number), Pres. (Presence), Color/Tex. (Color/Texture), Num. (Number), Shape (Shape), Posture (Posture), Pos. (Position), Abstract. (Abstract Knowledge), Concrete. (Concrete Knowledge), Expert (Expertise), Act. (Activity), Rel. (Relationships). $\uparrow$: higher is better, $\downarrow$: lower is better.}
    \vspace{0.3em}
    \setlength{\tabcolsep}{3pt}
    \renewcommand{\arraystretch}{1.5}
    \resizebox{\textwidth}{!}{%
        \begin{tabular}{lccccccccccccc}
            \specialrule{1.5pt}{0pt}{0pt}
            \rowcolor[HTML]{EFEFEF}
             Method & Char/Num & Pres. & Color/Tex & Num. & Shape & Posture & Pos. & Abstract. & Concrete. & Expert. & Act. & \multicolumn{1}{c}{Rel.} &  Avg. RA$\uparrow$\\
            \hline
            Random &25 &25 &25 &25 &25 &25 &25 &25 &25 &25 &25 &25 &25\\
            \hline
            Phi-3-vision (4B)~\cite{abdin2024phi3} &62.50 &59.09 &58.33 &37.50 &70.83 &33.33 &31.82 &54.55 &66.67 &41.94 &58.33 &50.00 &52.33\\
            Bunny-Llama-3-8B-V~\cite{he2024efficient} &55.00 &63.64 &54.17 &37.50 &79.17 &62.50 &54.55 &72.73 &85.71 &48.39 &75.00 &50.00 &60.67\\
            Idefics2-8B~\cite{laurençon2024matters} &57.50 &59.09 &54.17 &50.00 &79.17 &41.67 &27.27 &77.27 &76.19 &45.16 &75.00 &40.91 &56.67 \\
            \specialrule{1.5pt}{0pt}{0pt}\\
             \specialrule{1.5pt}{0pt}{0pt}
            \rowcolor[HTML]{EFEFEF}
             Method & Char/Num & Pres. & Color/Tex & Num. & Shape & Posture & Pos. & Abstract. & Concrete. & Expert. & Act. & \multicolumn{1}{c}{Rel.} & Avg. MR $\downarrow$\\
            \hline
            Random &50 &50 &50 &50 &50 &50 &50 &50 &50 &50 &50 &50 &50\\
            \hline
            Phi-3-vision (4B)~\cite{abdin2024phi3} &19.35 &18.75 &26.32 &43.75 &19.05 &55.56 &58.82 &40.00 &22.22 &48.00 &33.33 &31.25 &34.03\\
            Bunny-Llama-3-8B-V~\cite{he2024efficient} &15.38 &22.22 &18.75 &40.00 &5.00 &28.57 &29.41 &23.81 &10.00 &40.00 &10.00 &26.67 &22.22\\
            Idefics2-8B~\cite{laurençon2024matters} &23.33 &27.78 &23.53 &20.00 &13.64 &50.00 &40.00 &22.73 &11.11 &41.67 &14.29 &40.00 &26.72 \\
            \specialrule{1.5pt}{0pt}{0pt}
        \end{tabular}
    }
    \label{tab:with_image}
\end{table*}

\begin{table*}
    \centering
    \caption{Performance of MLLMs without images as input on the MMVU test set. $^\mathsection$We evaluate the officially released checkpoint by ourselves. Abbreviations: Char/Num. (Character/Number), Pres. (Presence), Color/Tex. (Color/Texture), Num. (Number), Shape (Shape), Posture (Posture), Pos. (Position), Abstract. (Abstract Knowledge), Concrete. (Concrete Knowledge), Expert (Expertise), Act. (Activity), Rel. (Relationships). $\uparrow$: higher is better, $\downarrow$: lower is better.}
    \vspace{0.3em}
    \setlength{\tabcolsep}{3pt}
    \renewcommand{\arraystretch}{1.5}
    \resizebox{\textwidth}{!}{%
        \begin{tabular}{lccccccccccccc}
            \specialrule{1.5pt}{0pt}{0pt}
            \rowcolor[HTML]{EFEFEF}
             Method & Char/Num & Pres. & Color/Tex & Num. & Shape & Posture & Pos. & Abstract. & Concrete. & Expert. & Act. & \multicolumn{1}{c}{Rel.} &  Avg. RA$\uparrow$\\
            \hline
            Random &25 &25 &25 &25 &25 &25 &25 &25 &25 &25 &25 &25 &25\\
            \hline
            Phi-3-vision (4B)~\cite{abdin2024phi3} &15.0 &13.64 &20.83 &0.0 &12.5 &8.33 &9.09 &27.27 &19.05 &16.13 &12.5 &13.64 &14.00 \\
            Bunny-Llama-3-8B-V~\cite{he2024efficient} &22.5 &13.64 &16.67 &4.17 &4.17 &16.67 &22.73 &22.73 &28.57 &19.35 &8.33 &27.27 &17.33 \\
            Idefics2-8B~\cite{laurençon2024matters} &15.0 &4.55 &8.33 &4.17 &8.33 &4.17 &0.0 &31.82 &23.81 &3.23 &0.0 &13.64 &9.76\\
            \specialrule{1.5pt}{0pt}{0pt}\\
             \specialrule{1.5pt}{0pt}{0pt}
            \rowcolor[HTML]{EFEFEF}
             Method & Char/Num & Pres. & Color/Tex & Num. & Shape & Posture & Pos. & Abstract. & Concrete. & Expert. & Act. & \multicolumn{1}{c}{Rel.} & Avg. MR $\downarrow$\\
            \hline
            Random &50 &50 &50 &50 &50 &50 &50 &50 &50 &50 &50 &50 &50\\
            \hline
            Phi-3-vision (4B)~\cite{abdin2024phi3} &72.73 &72.73 &58.33 &100.0 &66.67 &75.0 &81.82 &57.14 &60.0 &72.22 &62.5 &72.73 &69.34 \\
            Bunny-Llama-3-8B-V~\cite{he2024efficient} &64.0 &75.0 &63.64 &83.33 &91.67 &69.23 &44.44 &64.29 &50.0 &70.0 &80.0 &57.14 &67.09\\
            Idefics2-8B~\cite{laurençon2024matters} &77.78 &90.91 &80.0 &85.71 &85.71 &92.31 &100.0 &56.25 &61.54 &95.24 &100.0 &80.0 &82.10\\
            \specialrule{1.5pt}{0pt}{0pt}
        \end{tabular}
    }
    \label{tab:no_image}
\end{table*}

\section{MMVU-Train Data Construction Preliminary Experiment}
\subsection{Notes}
\begin{figure}[!t]
\centering
\includegraphics[width=0.8\linewidth]{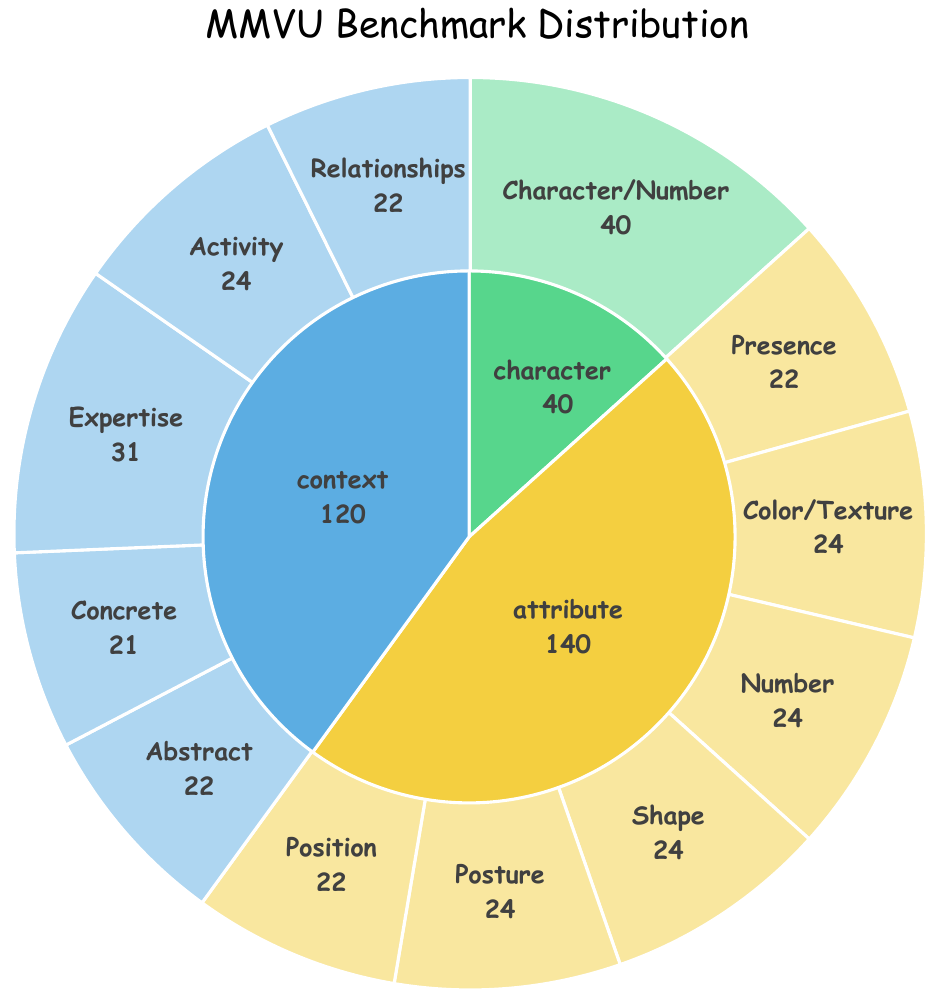}
\caption{Distribution of MMVU benchmark.}
\label{fig: distribution of MMR benchmark}
\end{figure}

\begin{enumerate}
    \item \textit{It is important to note that, to reduce the validation period, the test set used in the preparatory experiments does not constitute the full MMVU test set. The distribution of this test set, depicted in Fig.~\ref{fig: distribution of MMR benchmark}, closely resembles that of the final test set, with the primary difference being the scaling of each category. Consequently, the phenomena observed in the preparatory experiments on this subset are consistent with those observed using the full MMVU test set.}
    \item \textit{It should be noted that our data construction process is not carried out in multiple steps; rather, it is implicitly performed in a single step.}
\end{enumerate}

\subsection{Preliminary Testing Data Distribution.}
We have collected images that matched 12 subcategory themes from the web and some test sets, such as Flickr-30k~\cite{young2014image}, and facilitated setting misleading prompts. We meticulously design questions and standard answers based solely on the image information without relying on the original captions. 
The distribution of the subset of the MMVU test set is shown in Fig.~\ref{fig: distribution of MMR benchmark}. It categorizes questions into three main types: "context" (120 questions), ``character'' (40 questions), and ``attribute'' (140 questions). The ``context'' category includes subtypes such as relationships (22), activity (24), expertise (31), concrete (21), and abstract (22) questions. The ``character'' category comprises character/number questions (40). The ``attribute'' category encompasses presence (22), color/texture (24), number (24), shape (24), posture (24), and position (22) questions. This distribution highlights a comprehensive and balanced approach, ensuring a well-rounded benchmark with detailed subcomponents across various dimensions of question types.

\subsection{Training Details}
We employed a two-stage training strategy with specific hyperparameters for pre-training and fine-tuning, as detailed in Tab.~\ref{tab: parameters}. For testing, we used greedy search to ensure reproducibility.
\begin{table*}[h]
\centering
\caption{Hyperparameters.}
\vspace{0.3em}
\renewcommand{\arraystretch}{1.5}
\resizebox{\textwidth}{!}{\begin{tabular}{lcccccccc}
\specialrule{1.5pt}{0pt}{0pt}
\rowcolor[HTML]{EFEFEF}
 & batch size & lr & lr schedule & lr warmup ratio & weight decay & epoch & optimizer & DeepSpeed stage \\ \midrule
Pretrain & 256 & 1e-5 & cosine decay & 0.03 & 0 & 1 & AdamW & 2 \\ 
Finetune & 128 & 2e-4 & cosine decay & 0.03 & 0 & 1 & AdamW & 3 \\ \specialrule{1.5pt}{0pt}{0pt}
\end{tabular}}
\label{tab: parameters}
\end{table*}

\subsection{Discussion 1: Can We Simply Introduce Negative Samples to Train MLLMs?}
\label{Can we simply introduce negative samples to train MLLMs?}
We observed that many MLLMs could understand the content but struggled to answer questions. We hypothesized that the insufficiency of negative samples might be the cause. To address this, we generated additional negative samples and incorporated them into the training process. We sampled 7k and 24k samples from LLaVA and used GPT-4V to generate the negative samples. 

\paragraph{Prompt.} We employ two types of prompts for this task \textit{version 0}, which restricts only the number of problems, and \textit{version 1}, which restricts both the type and number of problems. Their common prompt is partially displayed in Fig.~\ref{fig:prompt_common}, while their respective different prompts are displayed in Fig.~\ref{fig:prompt_version0} and Fig.~\ref{fig:prompt_version1}, respectively. 
\begin{figure*}[h!]
    \centering
    \begin{tcolorbox}[width=\textwidth]
        \small
        \textbf{STEP-1: Parse and extract information based on the Input JSON and Information Topics:}
        \begin{enumerate}[leftmargin=5mm]
            \setlength{\itemsep}{4pt}
            \item Determine if there is any very clear text or numbers (You must make sure the identification is right) in the image. If yes, extract the corresponding text or numbers along with their position.
            \item Determine if there is any object in the image, including the properties of that object, such as name/category/color/texture/shape/pose (The properties of an object may not be unique, such as color, it may be a mixture of different colors), position, and quantity (You must make sure the identification is right).
            \item Determine if there is any person in the image, including attributes such as name, position, gender, and facial expression (You must make sure the identification is right. You must differentiate between a portrait in a photo or poster and a real person, don't confuse them.).
            \item Understand the events occurring in the image at a global level, and assess whether they relate to culture, emotions, or knowledge (You must make sure the identification is right).
            \item Understand the relationships between objects within the image, such as relative positions or hierarchical relationships (You must make sure the identification is right).
            \item Understand the relationships between people and objects, such as someone riding a bike or a person's actions (You must make sure the identification is right).
            \item If there are two or more people present, extract activities they are engaged in.
        \end{enumerate}
        
    \end{tcolorbox}
    \caption{Common Prompt for information extraction.}
    \label{fig:prompt_common}
\end{figure*}
\begin{figure*}[h!]
    \centering
    \begin{tcolorbox}[width=\textwidth]
        \small
        \textbf{STEP-2: Generate questions and corresponding correct answers based on the extracted information in STEP-1, Input Image, and Question Types (Do not output):}\\
        \textbf{Question Types:}
        \begin{enumerate}[leftmargin=5mm]
            \setlength{\itemsep}{4pt}
            \item If there is no text/number recognized or you do not make sure the text/number is correct, do not ask this type of question. If text is present, inquire about modifications to characters or numbers: Generate questions involving minor alterations, such as reversing the order of digits (modifying a number) or replacing a character.
            \item If characters or numbers are present in the image, modify one letter of a word or one digit of a number, then ask if the modified character or number exists, and add a distractor sentence to make the question more challenging. Otherwise, do not ask this type of question.
            \item Ask if certain objects exist: Replace the nouns of objects in the image with visually similar but nonexistent objects, then ask if the model identifies the object.
            \item Ask if objects exist and add a distractor sentence: Add a distractor sentence to questions based on modifying characters or numbers to make the question more challenging.
            \item If objects have attributes such as names of people, colors of objects, counts of objects, or positions, ask modified questions about object attributes and add a distractor sentence: Add a distractor sentence to questions based on modifying characters or numbers to guide the model to make incorrect inferences or answers.
            \item Ask global understanding questions: Understand the broader context, culture, emotions, and knowledge depicted in the image. Pose questions that differ slightly from true understanding but are very similar.
            \item Ask global understanding questions and add a distractor sentence: Add a distractor sentence to questions based on modifying characters or numbers to make the question more challenging.
            \item Ask local understanding questions: Based on the extracted relationships between objects, between people and objects, or between people, actions, or relative positions, pose local understanding challenges (negative samples).
            \item Ask local understanding questions and add a distractor sentence: Add a distractor sentence to questions based on modifying characters or numbers to guide the model to make incorrect inferences or answers.
        \end{enumerate}

        \textbf{Requirements:}
        \begin{enumerate}[leftmargin=5mm]
             \item Generate 7-10 questions.
             \item Ask directly, without adding any information such as "in the conversation," "mentioned in the conversation," or "in the image."
             \item Answer directly, with explanation, succinctly, with responses not exceeding 30 characters.
             \item Questions must be real and answerable. Answers must be correct, definitely,and not speculative.
             \item Do not generate questions that the answer is "uncertain".
             \item The format of the generated questions and answers is: 'conversations': [{'from': 'human', 'value': 'Question'}, {'from': 'gpt', 'value': 'Answer'}, {'from': 'human', 'value': 'Question'}, {'from': 'gpt', 'value': 'Answer'}, ..., {'from': 'human', 'value': 'Question'}, {'from': 'gpt', 'value': 'Answer'}]
        \end{enumerate}
        \textbf{STEP-3: Output JSON sample}
    \end{tcolorbox}
    \caption{GPT4 prompt for \textit{version 0}.}
    \label{fig:prompt_version0}
\end{figure*}
\begin{figure*}[h!]
    \centering
    \begin{tcolorbox}[width=\textwidth]
        \small
        \textbf{STEP-2: Generate questions and corresponding correct answers based on the extracted information in STEP-1, Input Image, and Question Types (Do not output):}\\
        \textbf{Question Types:}
        \begin{enumerate}[leftmargin=10mm]
        \item \textbf{Character:}
            \begin{enumerate}[leftmargin=5mm]
                \setlength{\itemsep}{4pt}
                \item If no text/number is recognized or the text/number recognized is not sure, do not ask questions about text/number. If text exists, inquire about modifications to characters or numbers: Generate questions involving minor alterations, such as reversing the order of digits (modifying a number) or replacing a character.
                \item If there are characters or numbers in the image, modify one letter or one digit, then ask if the modified character or number exists, and add a distractor sentence to make the question more challenging. Otherwise, do not ask this type of question.
            \end{enumerate}
        \item \textbf{Semantic:}
                \begin{enumerate}[leftmargin=5mm]
                \setlength{\itemsep}{4pt}
                \item Ask if certain objects exist: Replace the nouns of objects in the image with visually similar but nonexistent objects, then ask if the model identifies the object.
                \item  Ask if objects exist and add a distractor sentence: Add a distractor sentence to questions based on modifying characters or numbers to make the question more challenging.
                \item  If objects have attributes such as names of people, colors of objects, counts of objects, or positions, ask modified questions about object attributes and add a distractor sentence: Add a distractor sentence to questions based on modifying characters or numbers to guide the model to make incorrect inferences or answers.
            \end{enumerate}
        \item \textbf{Understanding:}
                \begin{enumerate}[leftmargin=5mm]
                \setlength{\itemsep}{4pt}
                \item Ask global understanding questions: Understand the broader context, culture, emotions, and knowledge depicted in the image. Pose questions that differ slightly from true understanding but are very similar.
                \item  Ask global understanding questions and add a distractor sentence: Add a distractor sentence to questions based on modifying characters or numbers to make the question more challenging.
                \item Ask local understanding questions: Based on the extracted relationships between objects, between people and objects, or between people, actions, or relative positions, pose local understanding challenges (negative samples).
                \item Ask local understanding questions and add a distractor sentence: Add a distractor sentence to questions based on modifying characters or numbers to guide the model to make incorrect inferences or answers.
            \end{enumerate}
        \end{enumerate}
            
        \textbf{Requirements:}
        \begin{enumerate}[leftmargin=5mm]
             \item Generate 2 character-type questions, 4 semantic-type questions, and 2 understanding-type questions.
             \item Ask directly, without adding any information such as "in the conversation," "mentioned in the conversation," or "in the image."
             \item Answer directly, with explanation, succinctly, with responses not exceeding 30 characters.
             \item Questions must be real and answerable. Answers must be correct, definitely,and not speculative.
             \item Do not generate questions that the answer is "uncertain".
             \item The format of the generated questions and answers is: 'conversations': [{'from': 'human', 'value': 'Question'}, {'from': 'gpt', 'value': 'Answer'}, {'from': 'human', 'value': 'Question'}, {'from': 'gpt', 'value': 'Answer'}, ..., {'from': 'human', 'value': 'Question'}, {'from': 'gpt', 'value': 'Answer'}]
        \end{enumerate}
        \textbf{STEP-3: Output JSON sample}
    \end{tcolorbox}
    \caption{GPT4 Prompt for \textit{version 1}.}
    \label{fig:prompt_version1}
\end{figure*}

\subsubsection{Training and Results of the \textit{Version 0}.}
\begin{figure*}[!t]
\centering
\includegraphics[width=0.8\linewidth]{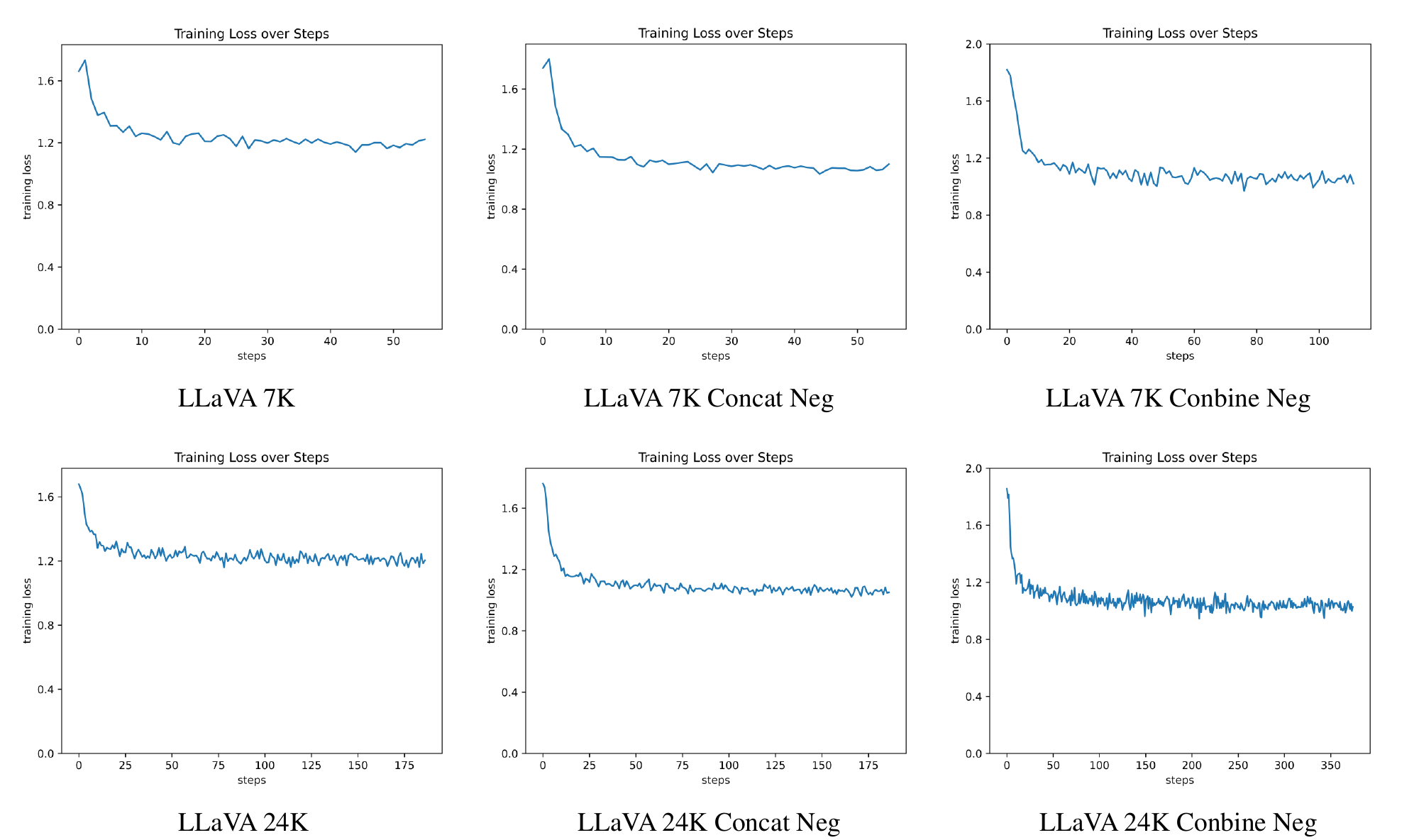}
\caption{Visualization of fine-tuning loss of different data composition (version 0).}
\label{fine-tuning loss}
\end{figure*}

We train MLLMs using the sampled data and synthetically generated data. The hyper-parameters used for training remain consistent with those outlined in Tab.~\ref{tab: parameters}. We first explore the model's performance using different data ratios: 1) sampled data alone, 2) sampled data concatenated with generated data, and 3) sampled data combination with generated data. The loss curves for the fine-tuning process are illustrated in Fig.~\ref{fine-tuning loss}. The loss curves in Fig.~\ref{fine-tuning loss} reveal that incorporating generated negative data, whether through concatenation or combination, leads to a further reduction in model loss compared to training solely on sampled data. Notably, the lowest loss is achieved when utilizing a combination of both sampled and generated data. 

Tab.~\ref{tab:appendix_promptv0_mma} and Tab.~\ref{tab:appendix_promptv0} present the model's performance on the MMVU test set and other benchmarks across the different data compositions. As expected, training exclusively on sampled data yields the lowest overall performance. Interestingly, while splicing sampled data with generated negative samples leads to a performance decrease on the generic benchmark, it demonstrates improvements on certain phantom benchmarks.
The most effective approach proved to be combining the original sampled data with generated negative samples. This strategy consistently enhances performance across most benchmarks, often by a significant margin.

\begin{table*}
    \centering
    \caption{Ablation about only introducing negative samples on the MMVU test set. Concat. and Neg. denotes concatenate and negative samples, respectively.}
    \vspace{0.3em}
    \setlength{\tabcolsep}{3pt}
    \renewcommand{\arraystretch}{1.5}
    \resizebox{\textwidth}{!}{%
        \begin{tabular}{lccccccccccccc}
            \specialrule{1.5pt}{0pt}{0pt}
            \rowcolor[HTML]{EFEFEF}
             Method & Char/Num & Pres. & Color/Tex & Num. & Shape & Stance & Pos. & Abstract. & Concrete. & Expert. & Act. & \multicolumn{1}{c}{Rel.} & Avg. RA$\uparrow$\\
            \hline
            LLaVA 7K & 2.50 & 0.00 & 4.17 & 0.00 & 0.00 & 0.00 & 4.55 & 0.00 & 4.76 & 0.00 & 0.00 & 4.55 & 1.67 \\
            LLaVA 7K Concat. Neg. & 2.50 & 0.00 & 4.17 & 4.17 & 20.83 & 0.00 & 4.55 & 0.00 & 0.00 & 3.23 & 4.17 & 4.55 & 4.00 \\
            LLaVA 7K Conbine Neg.  & 5.00 & 4.55 & 8.33 & 16.67 & 25.00 & 4.17 & 9.09 & 18.18 & 4.76 & 3.23 & 16.67 & 9.09 & 10.00 \\
            \hline
            LLaVA 24K & 17.50 & 22.73 & 20.83 & 12.50 & 45.83 & 16.67 & 13.64 & 50.00 & 28.57 & 29.03 & 45.83 & 31.82 & 27.33 \\
            LLaVA 24K Concat. Neg. & 20.00 & 13.64 & 8.33 & 16.67 & 58.33 & 12.50 & 4.55 & 40.91 & 19.05 & 9.68 & 25.00 & 31.82 & 21.33 \\
            LLaVA 24K Conbine Neg. & 27.50 & 27.27 & 25.00 & 20.83 & 54.17 & 16.67 & 9.09 & 45.45 & 38.10 & 25.81 & 54.17 & 36.36 & 31.33 \\
            \specialrule{1.5pt}{0pt}{0pt}
            \rowcolor[HTML]{EFEFEF}
            Method & Char/Num & Pres. & Color/Tex & Num. & Shape & Stance & Pos. & Abstract. & Concrete. & Expert. & Act. & \multicolumn{1}{c}{Rel.} & Avg. MR $\downarrow$\\
            \hline
            LLaVA 7K & 83.33 & 100.00 & 80.00 & 100.00 & 100.00 & 100.00 & 75.00 & 100.00 & 80.00 & 100.00 & 100.00 & 80.00 & 93.06 \\
            LLaVA 7K Concat. Neg.  & 88.89 & 100.00 & 66.67 & 87.50 & 44.44 & 100.00 & 66.67 & 100.00 & 100.00 & 92.31 & 75.00 & 75.00 & 84.42 \\
            LLaVA 7K Conbine Neg.  & 86.67 & 83.33 & 71.43 & 66.67 & 57.14 & 92.31 & 66.67 & 66.67 & 92.31 & 93.75 & 55.56 & 81.82 & 77.61 \\
            \hline
            LLaVA 24K & 72.00 & 54.55 & 68.75 & 75.00 & 38.89 & 77.78 & 62.50 & 45.00 & 68.42 & 62.50 & 38.89 & 46.15 & 59.41 \\
            LLaVA 24K Concat. Neg. & 52.94 & 62.50 & 85.71 & 63.64 & 26.32 & 80.00 & 87.50 & 50.00 & 77.78 & 83.33 & 53.85 & 50.00 & 63.01 \\
            LLaVA 24K Conbine Neg. & 59.26 & 50.00 & 60.00 & 58.33 & 31.58 & 78.95 & 77.78 & 47.37 & 60.00 & 68.00 & 35.00 & 42.86 & 55.45 \\
            \specialrule{1.5pt}{0pt}{0pt}
        \end{tabular}
    }
    \label{tab:appendix_promptv0_mma}
\end{table*}
\begin{table*}
    \centering
\caption{Ablation about only introducing negative samples on general benchmarks. Concat. and Neg. denotes concatenate and negative samples, respectively.}
\vspace{0.3em}
\renewcommand{\arraystretch}{1.5}
\resizebox{0.7\textwidth}{!}{\begin{tabular}{lccccc}
\specialrule{1.5pt}{0pt}{0pt}
\rowcolor[HTML]{EFEFEF}
{Dataset} & {MME$^\text{P}$} & {MME$^\text{C}$} &  MMB$^\text{T}$ & MMB$^\text{D}$ & POPE \\
\midrule
LLaVA 7K & 719.07 & 272.86 & 1.68 & 2.23 & 74.70\\
LLaVA 7K Concat. Neg. & 1095.19 & 267.5 & 0.50 & 1.03 & 82.69\\
LLaVA 7K Conbine Neg. & 883.14 & 287.14 & 4.76 & 4.90 & 82.99\\
\midrule
LLaVA 24K & 1039.62 & 265.36 & 41.48 & 42.53 & 82.51\\
LLaVA 24K Concat. Neg. & 1308.59 & 254.64 & 15.53 & 18.21 & 82.22\\
LLaVA 24K Conbine Neg. & 1174.62 & 270.35 & 49.61 & 53.78 & 84.70\\
\specialrule{1.5pt}{0pt}{0pt}
\end{tabular}}
    \label{tab:appendix_promptv0}
\end{table*}

\begin{figure*}[!t]
\centering
\includegraphics[width=0.8\linewidth]{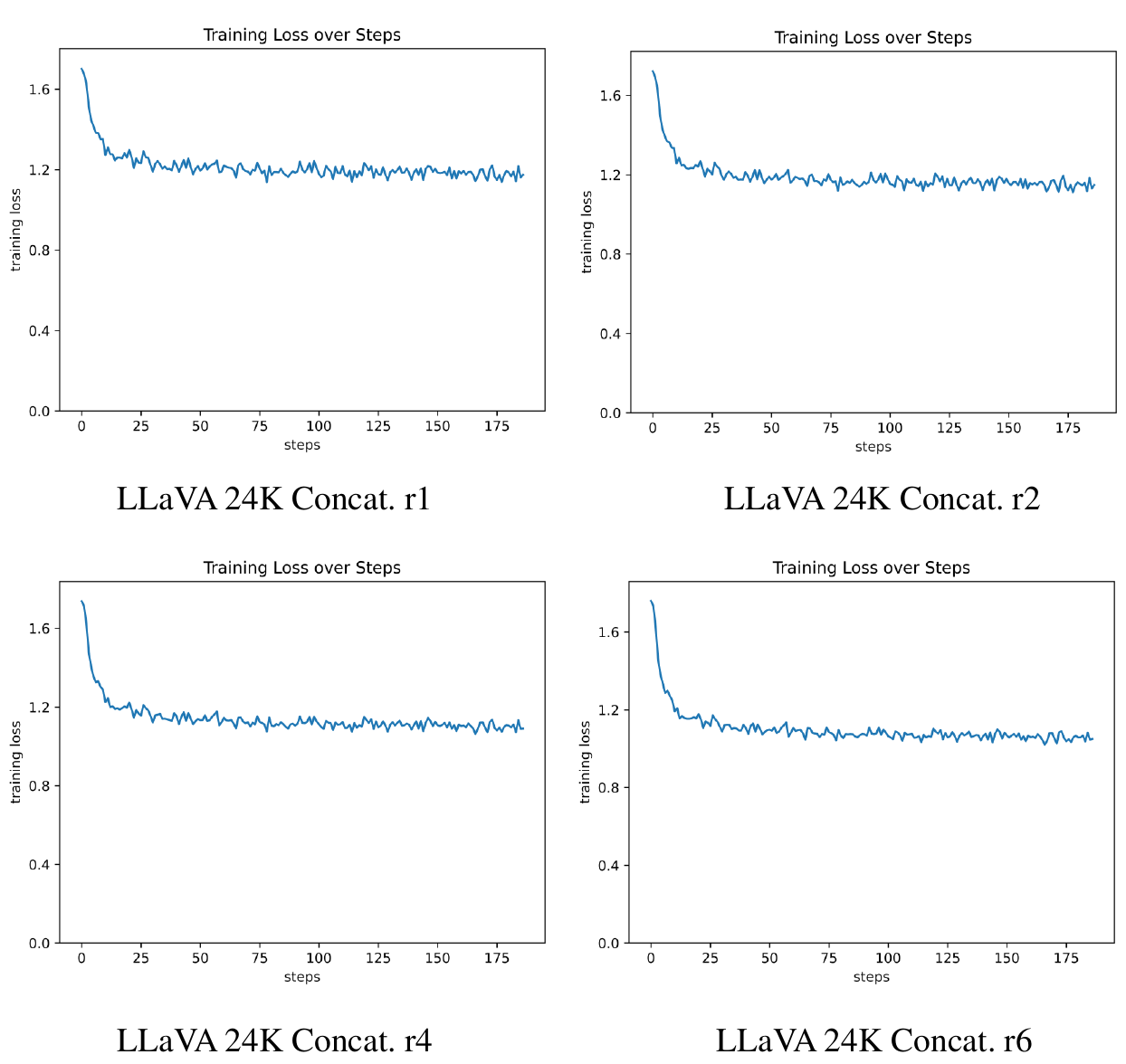}
\caption{Visualization of fine-tuning loss of different conversation rounds.}
\label{fig: appendix-Ablation about conversation rounds.}
\end{figure*}

\paragraph{Ablation about conversation rounds.} Our initial concatenation approach directly appended all conversation rounds of a negative sample to the original data. However, this might lead to performance degradation due to data imbalance. To investigate this, we conducted an ablation study on the 24k dataset, focusing on the number of conversation rounds included from negative samples.
Fig.~\ref{fig: appendix-Ablation about conversation rounds.} presents the model's training loss across different numbers of conversation rounds. Additionally, Tab.~\ref{tab:appendix_promptv0_cr_mma} and Tab.~\ref{tab:appendix_promptv0_cr} showcase the corresponding performance on the MMVU and generalized benchmarks, respectively. The experimental results indicate that using fewer concatenated conversation rounds does not significantly degrade model performance. However, a minor performance drop is still observed. We hypothesize that this drop stems from the increased variability within each training sample when fewer rounds are used. This variability might make it more challenging for the model to learn consistent patterns and generalize effectively.

\begin{table*}
    \centering
\caption{Ablation about only introducing negative samples on general benchmarks. Concat. and Neg. denotes concatenate and negative samples, respectively.}
\vspace{0.3em}
\renewcommand{\arraystretch}{1.5}
\resizebox{0.7\textwidth}{!}{\begin{tabular}{lccccc}
\specialrule{1.5pt}{0pt}{0pt}
\rowcolor[HTML]{EFEFEF}
{Dataset} & {MME$^\text{P}$} & {MME$^\text{C}$} &  MMB$^\text{T}$ & MMB$^\text{D}$ & POPE \\
\midrule
LLaVA 24K & 1039.62 & 265.36 & 41.48 & 42.53 & 82.51\\
LLaVA 24K Concat. Neg. (r1) & 1304.77 & 268.21 & 40.87 & 41.87 & 81.65\\
LLaVA 24K Concat. Neg. (r2) & 1336.41 & 258.57 & 24.74 & 18.21 & 80.77\\
LLaVA 24K Concat. Neg. (r4) & 1304.25 & 259.29 & 25.00 & 28.52 & 82.42\\
LLaVA 24K Concat. Neg. (all )& 1308.59 & 254.64 & 15.53 & 18.21 & 82.22\\
\specialrule{1.5pt}{0pt}{0pt}
\end{tabular}}
    \label{tab:appendix_promptv0_cr}
\end{table*}

\begin{table*}
    \centering
    \caption{Ablation about only introducing negative samples on the MMVU test set. Concat. and Neg. denotes concatenate and negative samples, respectively.}
    \vspace{0.3em}
    \setlength{\tabcolsep}{3pt}
    \renewcommand{\arraystretch}{1.5}
    \resizebox{\textwidth}{!}{%
        \begin{tabular}{lccccccccccccc}
            \specialrule{1.5pt}{0pt}{0pt}
            \rowcolor[HTML]{EFEFEF}
             Method & Char/Num & Pres. & Color/Tex & Num. & Shape & Stance & Pos. & Abstract. & Concrete. & Expert. & Act. & \multicolumn{1}{c}{Rel.} & Avg. RA$\uparrow$\\
            \hline
            LLaVA 24K & 17.50 & 22.73 & 20.83 & 12.50 & 45.83 & 16.67 & 13.64 & 50.00 & 28.57 & 29.03 & 45.83 & 31.82 & 27.33 \\
            LLaVA 24K Concat. Neg. (r1) & 22.50 & 22.73 & 16.67 & 16.67 & 45.83 & 20.83 & 9.09 & 45.45 & 23.81 & 22.58 & 45.83 & 27.27 & 26.33 \\
            LLaVA 24K Concat. Neg. (r2)  & 17.50 & 13.64 & 8.33 & 12.50 & 58.33 & 16.67 & 9.09 & 45.45 & 38.10 & 16.13 & 37.50 & 31.82 & 24.67 \\
            LLaVA 24K Concat. Neg. (r4)  & 17.50 & 9.09 & 20.83 & 16.67 & 70.83 & 16.67 & 13.64 & 45.45 & 33.33 & 19.35 & 45.83 & 36.36 & 28.00 \\
            LLaVA 24K Concat. Neg. (all)  & 20.00 & 13.64 & 8.33 & 16.67 & 58.33 & 12.50 & 4.55 & 40.91 & 19.05 & 9.68 & 25.00 & 31.82 & 21.33 \\
            \specialrule{1.5pt}{0pt}{0pt}
            \rowcolor[HTML]{EFEFEF}
             Method & Char/Num & Pres. & Color/Tex & Num. & Shape & Stance & Pos. & Abstract. & Concrete. & Expert. & Act. & \multicolumn{1}{c}{Rel.} & Avg. MR $\downarrow$\\
            \hline
            LLaVA 24K & 72.00 & 54.55 & 68.75 & 75.00 & 38.89 & 77.78 & 62.50 & 45.00 & 68.42 & 62.50 & 38.89 & 46.15 & 59.41 \\
            LLaVA 24K Concat. Neg. (r1) & 65.38 & 61.54 & 73.33 & 69.23 & 42.11 & 73.68 & 80.00 & 47.37 & 73.68 & 72.00 & 38.89 & 53.85 & 62.20 \\
            LLaVA 24K Concat. Neg. (r2)  & 65.00 & 72.73 & 85.71 & 75.00 & 33.33 & 76.47 & 80.00 & 47.37 & 60.00 & 77.27 & 43.75 & 53.33 & 62.44 \\
            LLaVA 24K Concat. Neg. (r4)  & 63.16 & 80.00 & 66.67 & 60.00 & 19.05 & 75.00 & 66.67 & 47.37 & 65.00 & 71.43 & 42.11 & 42.86 & 56.48 \\
            LLaVA 24K Concat. Neg. (all)  & 52.94 & 62.50 & 85.71 & 63.64 & 26.32 & 80.00 & 87.50 & 50.00 & 77.78 & 83.33 & 53.85 & 50.00 & 63.01 \\
            \specialrule{1.5pt}{0pt}{0pt}
        \end{tabular}
    }
    \label{tab:appendix_promptv0_cr_mma}
\end{table*}

\subsubsection{Training and Results of the \textit{Version 1}.}
\begin{figure*}[!t]
\centering
\includegraphics[width=0.8\linewidth]{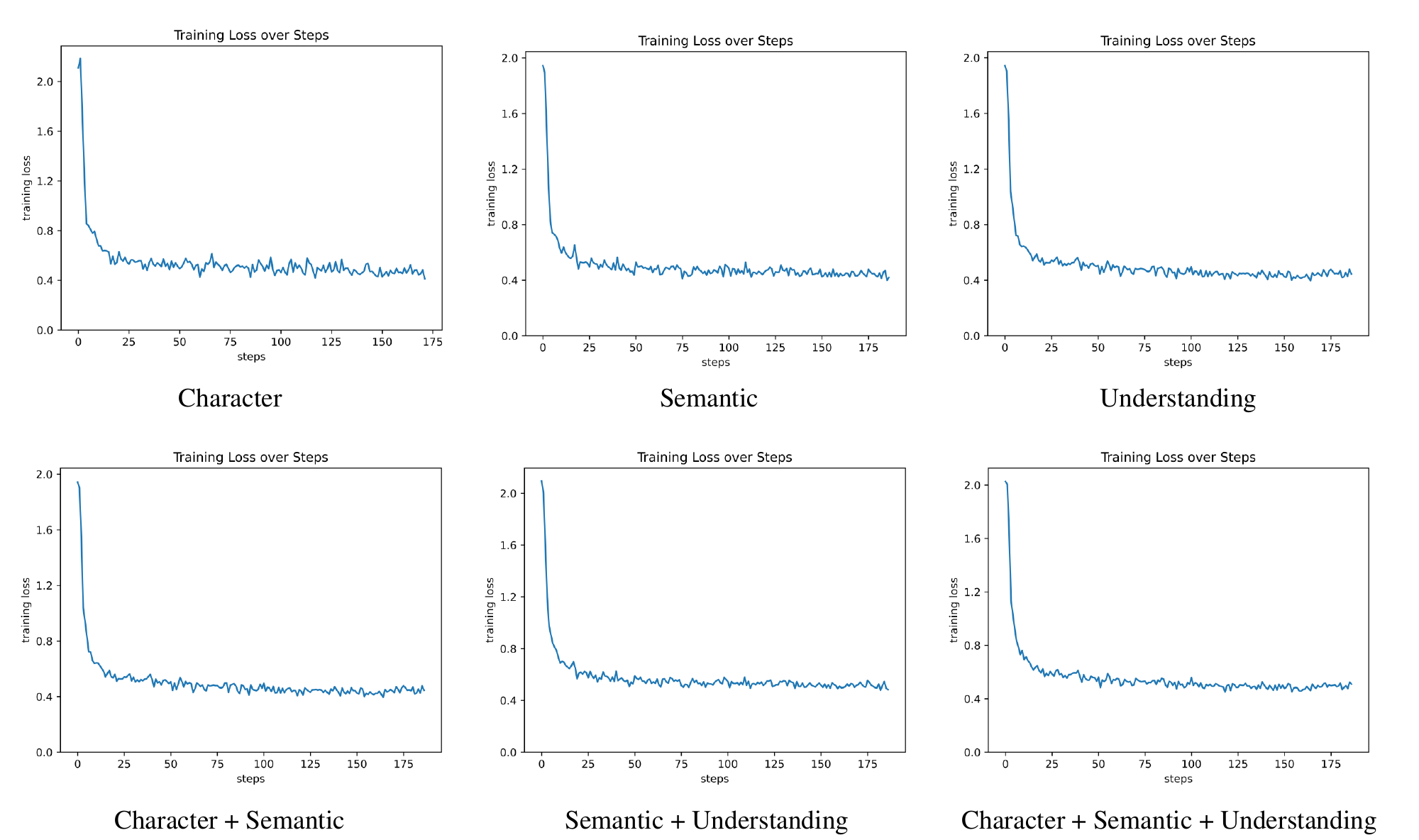}
\caption{Visualization of fine-tuning loss of different data composition (version 1).}
\label{tab: appendix-version1 loss}
\end{figure*}

To further investigate the model's performance under constraints on question types, we introduce a modified prompt and generate a new 24k dataset for experiments. Interestingly, despite prompting for negative samples, GPT-4V still produced a mix of positive and negative examples, with an approximate ratio of 1:13. Utilizing only this generated dataset, we have conducted experiments to analyze the impact of different problem types. Fig.~\ref{tab: appendix-version1 loss} illustrates the training loss, while Tab.~\ref{tab:appendix_promptv1} and Tab.~\ref{tab:appendix_promptv1_mma} present the corresponding performance results on benchmarks.

\begin{table*}
    \centering
\caption{Ablation about different question types.}
\vspace{0.3em}
\renewcommand{\arraystretch}{1.5}
\resizebox{.8\textwidth}{!}{\begin{tabular}{lccccc}
\specialrule{1.5pt}{0pt}{0pt}
\rowcolor[HTML]{EFEFEF}
{Dataset} & {MME$^\text{P}$} & {MME$^\text{C}$} &  MMB$^\text{T}$ & MMB$^\text{D}$ & POPE \\
\midrule
Character & 869.21 & 203.93 & 58.63 & 58.59 & 66.18 \\
Semantic & 1167.73 & 216.78 & 61.43 & 61.52 & 71.33 \\
Understanding & 918.38 & 219.64 & 60.15 & 59.88 & 67.93 \\
Character + Semantic & 963.81 & 225.36 & 59.98 & 60.05 & 68.84 \\
Semantic + Understanding & 1260.70 & 246.43 & 62.05 & 61.77 & 71.82 \\
Character + Semantic + Understanding & 1051.6 & 242.5 & 61.27 & 61.00 & 71.63\\
\specialrule{1.5pt}{0pt}{0pt}
\end{tabular}}
    \label{tab:appendix_promptv1}
\end{table*}
\begin{table*}
    \centering
    \caption{Ablation about only introducing negative samples on the MMVU test set. Concat. and Neg. denotes concatenate and negative samples, respectively.}
    \vspace{0.3em}
    \setlength{\tabcolsep}{3pt}
    \renewcommand{\arraystretch}{1.5}
    \resizebox{\textwidth}{!}{%
        \begin{tabular}{lccccccccccccc}
            \specialrule{1.5pt}{0pt}{0pt}
            \rowcolor[HTML]{EFEFEF}
             Method & Char/Num & Pres. & Color/Tex & Num. & Shape & Stance & Pos. & Abstract. & Concrete. & Expert. & Act. & \multicolumn{1}{c}{Rel.} & Avg. MA $\uparrow$\\
            \hline
            Character  & 35.00 & 40.91 & 20.83 & 20.83 & 62.50 & 33.33 & 22.73 & 59.09 & 33.33 & 25.81 & 50.00 & 31.82 & 36.00 \\
            Semantic & 27.50 & 31.82 & 33.33 & 33.33 & 79.17 & 29.17 & 18.18 & 45.45 & 47.62 & 35.48 & 58.33 & 45.45 & 39.67 \\
            Understanding & 47.50 & 40.91 & 29.17 & 33.33 & 66.67 & 41.67 & 31.82 & 54.55 & 47.62 & 38.71 & 58.33 & 40.91 & 44.33 \\
            Character + Semantic & 45.00 & 40.91 & 29.17 & 37.50 & 66.67 & 33.33 & 27.27 & 59.09 & 47.62 & 35.48 & 54.17 & 40.91 & 43.00 \\
            Semantic + Understanding  & 32.50 & 45.45 & 33.33 & 41.67 & 70.83 & 20.83 & 18.18 & 68.18 & 28.57 & 38.71 & 58.33 & 45.45 & 41.33\\
            Character + Semantic + Understanding & 42.50 & 36.36 & 29.17 & 37.50 & 70.83 & 29.17 & 27.27 & 54.55 & 42.86 & 35.48 & 58.33 & 45.45 & 42.33\\
            \specialrule{1.5pt}{0pt}{0pt}
            \rowcolor[HTML]{EFEFEF}
             Method & Char/Num & Pres. & Color/Tex & Num. & Shape & Stance & Pos. & Abstract. & Concrete. & Expert. & Act. & \multicolumn{1}{c}{Rel.} & Avg. MR $\downarrow$\\ 
            \hline
            Character  & 46.15 & 35.71 & 61.54 & 66.67 & 21.05 & 60.00 & 44.44 & 38.10 & 58.82 & 68.00 & 29.41 & 50.00 & 48.57 \\
            Semantic & 62.07 & 30.00 & 52.94 & 38.46 & 5.00 & 63.16 & 66.67 & 52.38 & 47.37 & 57.69 & 26.32 & 41.18 & 46.40 \\
            Understanding & 26.92 & 30.77 & 53.33 & 33.33 & 20.00 & 54.55 & 30.00 & 42.86 & 44.44 & 53.85 & 26.32 & 43.75 & 38.99 \\
            Character + Semantic & 30.77 & 35.71 & 58.82 & 25.00 & 23.81 & 57.89 & 45.45 & 38.10 & 47.37 & 57.69 & 27.78 & 47.06 & 41.63 \\
            Semantic + Understanding  & 51.85 & 23.08 & 52.94 & 28.57 & 15.00 & 72.22 & 63.64 & 31.82 & 68.42 & 55.56 & 30.00 & 44.44 & 45.13 \\
            Character + Semantic + Understanding & 37.04 & 33.33 & 56.25 & 25.00 & 19.05 & 63.16 & 50.00 & 45.45 & 50.00 & 54.17 & 26.32 & 44.44 & 42.27\\
            \specialrule{1.5pt}{0pt}{0pt}
        \end{tabular}
    }
    \label{tab:appendix_promptv1_mma}
\end{table*}
This experiment yielded several key insights: \textbf{1) Avoid rigidly specifying the number of question types, particularly for characteristics like characters.} Enforcing such constraints, especially when not all images contain those characteristics, can lead to generating nonsensical question-answer pairs and consequently, performance degradation. \textbf{2) Training solely on generated data can be surprisingly effective, but only when both positive and negative sample pairs are present.} This finding highlights the potential of synthetic data for training MLLMs. Notably, using only negative samples restricts the model from producing negative answers, emphasizing the importance of balanced datasets. Consequently, we shifted our focus to \textbf{generating paired positive and negative samples} to investigate model performance under this training paradigm further.

\subsection{Discussion 2: What Constitutes an Effective Pair of Positive and Negative Samples for Training?}
\begin{figure*}[!t]
\centering
\includegraphics[width=0.8\linewidth]{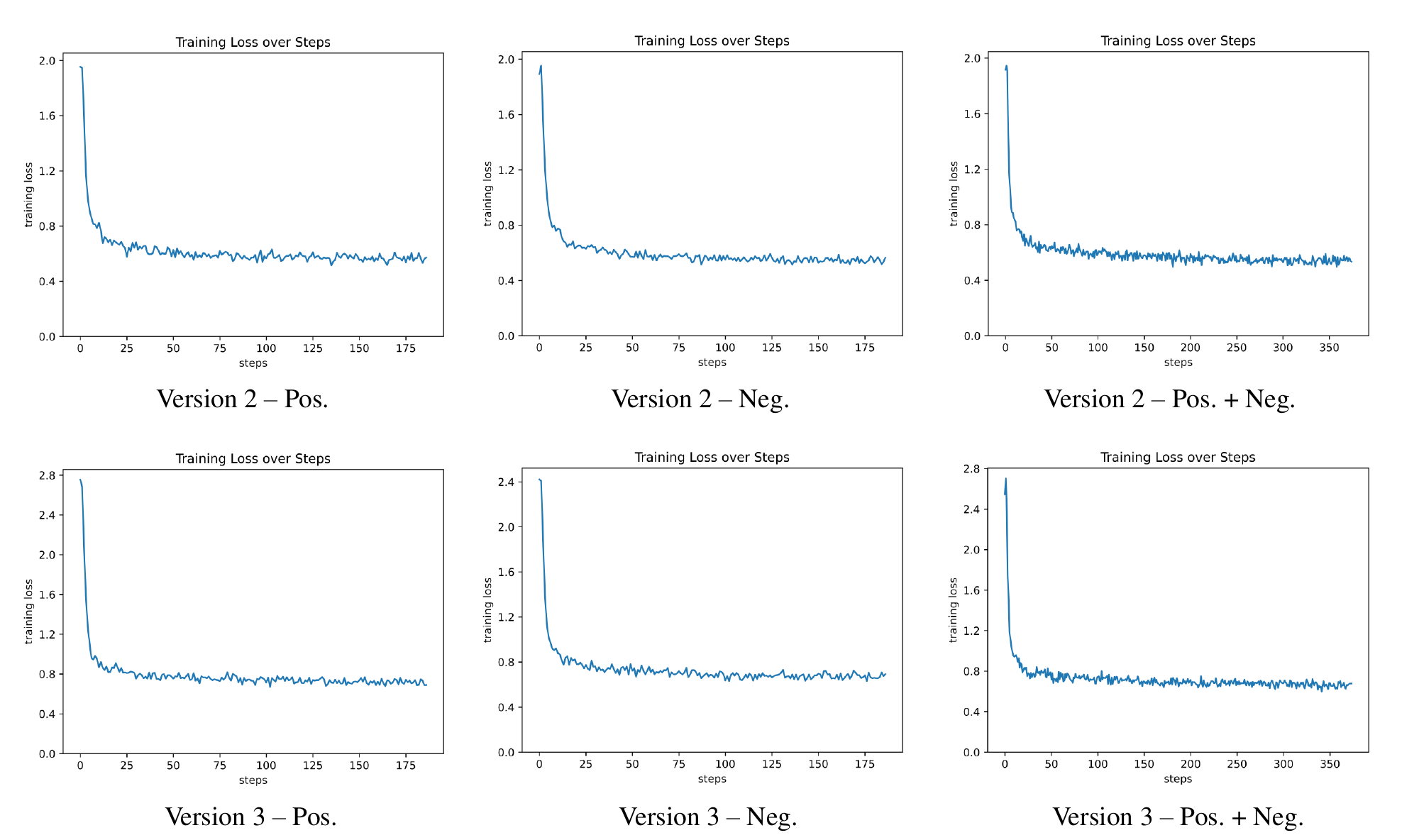}
\caption{Visualization of fine-tuning loss of different data compositions (version 2 and version 3).}
\label{fig: appendix-version2_3 loss}
\end{figure*}

Having established the benefit of incorporating both positive and negative samples during training, we now delve into effective construction methods for such pairs. We investigate two distinct approaches: \textbf{1) Explicit Positive/negative Prompting.} This method involves directly prompting the model to generate both positive and negative samples, with answers explicitly labeled as "yes" or "no." This approach offers straightforward control over the positive/negative balance in the training data (\textit{version 2}). \textbf{Confounding Answer Construction.}
This approach focuses on crafting questions that naturally lend themselves to multiple plausible answers, without explicitly forcing a "yes/no" response. This strategy aims to create a more nuanced and challenging training dataset, potentially enhancing the model's ability to discern subtle differences in meaning and context (\textit{version 3}). The GPT-4V prompts for version 2 and version 3 are shown in Fig.~\ref{fig:prompt_version2} and Fig.~\ref{fig:prompt_version3}, respectively. To evaluate the effectiveness of these two versions of prompts, we generated 24k positive-negative sample pairs using each version's prompt. We then incorporated these datasets into our training pipeline, maintaining consistent experimental settings with previous trials. Fig.~\ref{fig: appendix-version2_3 loss} presents the training loss curves observed for both approaches, while Tab.~\ref{tab:appendix_promptv2_3} and Tab.~\ref{tab:appendix_promptv2_3_mma} detail the corresponding performance results across various benchmarks. Our experimental results demonstrate that utilizing prompt version 3 to generate paired positive and negative samples yielded the strongest overall performance. Consequently, we adopted version 3 as our preferred method for generating training data in the main experiments.
 
 \begin{figure*}[h!]
    \centering
    \begin{tcolorbox}[width=\textwidth]
        \small
        \textbf{STEP-2: Generate questions and corresponding correct answers based on the extracted information in STEP-1, Input Image, and Question Types (Do not output):}\\
        \textbf{Question Types:}
        \begin{enumerate}[leftmargin=5mm]
            \setlength{\itemsep}{4pt}
            \item Ask if a certain object exists, the question is defined as ques-pos. Provide the correct answer to ques-pos as ans-pos. (2) Replace the object mentioned in the ques-pos and ans-pos with the visually similar nonexistent object. Ask if the model identifies the replaced object, the question is defined as ques-neg. Rhetorical questions are preferred. Provide the correct answer to ques-neg as ans-neg.
            \item Ask a question about the name/category/color/texture/shape/pose of an object in the image, the question is defined as ques-pos. Provide the correct answer to ques-pos as ans-pos. (2) Replace the name/category/color/texture/shape/pose of the object mentioned in the ques-pos. Ask a question about if the modified name/category/color/texture/shape/pose of an object is correct. Rhetorical questions are preferred. The new question is defined as ques-neg. Provide the correct answer to ques-neg as ans-neg. 
            \item Ask a question about the number/position of an object in the image, the question is defined as ques-pos. Provide the correct answer to ques-pos as ans-pos. (2) Replace the number/position of the object mentioned in the ques-pos with a similar number/position as interference. Ask a question about if the modified number/position of the object is correct. Rhetorical questions are preferred. The new question is defined as ques-neg. Provide the correct answer to ques-neg as ans-neg. 
            \item Ask a question about the topic of abstract-knowledge/concrete-knowledge/professional-knowledge in the image, the question is defined as ques-pos. Provide the correct answer to ques-pos as ans-pos. (2) Replace the knowledge related to the topic mentioned in the ques-pos with similar knowledge as interference. Ask a distractor question about if the modified knowledge related to the topic is correct. Rhetorical questions are preferred. The new question is defined as ques-neg. Provide the correct answer to ques-neg as ans-neg. 
            - Candidate topics of abstract knowledge: emotions, aesthetics, connotations, symbols, culture, news, allusions, legends, common sense, functions, and so on.
            - Candidate topics of concrete knowledge: landmarks, celebrities, well-known objects, and so on.
            - Candidate topics of professional knowledge: specific knowledge in various vertical fields, industries, disciplines, and so on.
            \item Ask a question about the activity of an object and/or interaction/relationship between objects or persons in the image, the question is defined as ques-pos. Provide the correct answer to ques-pos as ans-pos. (2) Replace the activity/interaction/relationship mentioned in the ques-pos with a similar activity/interaction/relationship as interference. Ask a distractor question about if the modified activity/interaction/relationship is correct. Rhetorical questions are preferred. The new question is defined as ques-neg. Provide the correct answer to ques-neg as ans-neg. 
            \item NOTE: Pose questions that differ slightly from true understanding but are very similar, or add a distractor sentence to questions, so that the model is guided to make incorrect inferences or answers.
            \item NOTE: Pose questions that differ slightly from true understanding but are very similar, or add a distractor sentence to questions, so that the model is guided to make incorrect inferences or answers.
        \end{enumerate}

        \textbf{Requirements:}
        \begin{enumerate}[leftmargin=5mm]
             \item Generate 7-10 questions.
             \item Ask directly, without adding any information such as "in the conversation," "mentioned in the conversation," or "in the image."
             \item Answer directly, with explanation, succinctly, with responses not exceeding 30 characters.
             \item Questions must be real and answerable. Answers must be correct, definitely, and not speculative.
             \item Do not generate questions that the answer is "uncertain".
             \item The format of the generated questions and answers is: {'id': '{json['id']}', 'image': '{json['image']}', 'conversations-pos': [{'from': 'human', 'value': 'ques-pos'}, {'from': 'gpt', 'value': 'ans-pos'},..., {'from': 'human', 'value': 'ques-pos'}, {'from': 'gpt', 'value': 'ans-pos'}], 'conversations-neg': [{'from': 'human', 'value': 'ques-neg '}, {'from': 'gpt', 'value': 'ans-neg'},..., {'from': 'human', 'value': 'ques-neg'}, {'from': 'gpt', 'value': 'ans-neg'}]}
        \end{enumerate}
        \textbf{STEP-3: Output JSON sample}
    \end{tcolorbox}
    \caption{GPT4 prompt for \textit{version 2}.}
    \label{fig:prompt_version2}
\end{figure*}
 \begin{figure*}[h!]
    \centering
    \begin{tcolorbox}[width=\textwidth]
        \small
        \textbf{STEP-2: Generate questions and corresponding correct answers based on the extracted information in STEP-1, Input Image, and Question-Answer Types:}

        \textbf{Question-Answer Types:}
        \begin{enumerate}[leftmargin=10mm]
            \item \textbf{Types-1:} (1) Ask if a certain object exists, the question is defined as ques-pos. (2) Operation-pos (3) Replace the object mentioned in the ques-pos with the visually similar nonexistent object. Ask if the model identifies the replaced object, the question is defined as ques-neg. Rhetorical questions are preferred. (4) Operation-neg
            \item \textbf{Types-2:} (1) Ask a question about the name/category/color/texture/shape/pose of an object in the image, the question is defined as ques-pos. (2) Operation-pos (3) Replace the name/category/color/texture/shape/pose of the object mentioned in the ques-pos. Ask a question about whether the modified name/category/color/texture/shape/pose of an object is correct. Rhetorical questions are preferred. The new question is defined as ques-neg. (4) Operation-neg
            \item \textbf{Types-3:}(1) Ask a question about the number/position of an object in the image, the question is defined as ques-pos. (2) Operation-pos (3) Replace the number/position of the object mentioned in the ques-pos with a similar number/position as interference. Ask a question about if the modified number/position of the object is correct. Rhetorical questions are preferred. The new question is defined as ques-neg. (4) Operation-neg
            \item \textbf{Types-4:}(1) Ask a question about the topic of abstract-knowledge/concrete-knowledge/professional-knowledge in the image, the question is defined as ques-pos. (2) Operation-pos (3) Replace the knowledge related to the topic mentioned in the ques-pos with similar knowledge as interference. Ask a distractor question about if the modified knowledge related to the topic is correct. Rhetorical questions are preferred. The new question is defined as ques-neg. (4) Operation-neg
                \begin{itemize}
                    \item Candidate topics of abstract knowledge: emotions, aesthetics, connotations, symbols, culture, news, allusions, legends, common sense, functions, and so on.
                    \item Candidate topics of concrete knowledge: landmarks, celebrities, well-known objects, and so on.
                    \item Candidate topics of professional knowledge: specific knowledge in various vertical fields, industries, disciplines, and so on.
                \end{itemize}
            \item \textbf{Types-5:}(1) Ask a question about the activity of an object and/or interaction/relationship between objects or persons in the image, the question is defined as ques-pos. (2) Operation-pos (3) Replace the activity/interaction/relationship mentioned in the ques-pos with a similar activity/interaction/relationship as interference. Ask a distractor question about if the modified activity/interaction/relationship is correct. Rhetorical questions are preferred. The new question is defined as ques-neg. (4) Operation-neg
        \end{enumerate}

        \textbf{Notes:}
        \begin{itemize}[leftmargin=5mm]
            \item All (2) Operation-pos is the same type: Design and provide four options based on ques-pos, including opt-pos-0: right answer with the correct reason, opt-pos-1: right answer with an incorrect reason, opt-pos-2: wrong answer with the correct reason, opt-pos-3: wrong answer with an incorrect reason. The opt-pos-0 is defined as ans-pos.
            \item All (4) Operation-neg is the same type: Design and provide four options based on ques-neg, including opt-neg-0: right answer with the correct reason, opt-neg-1: right answer with an incorrect reason, opt-neg-2: wrong answer with the correct reason, opt-neg-3: wrong answer with an incorrect reason. The opt-neg-0 is defined as ans-neg.
            \item The correct reason is the only one, while the incorrect reason can stem from various distortions and fabrications of facts, etc.
            \item Pose questions that differ slightly from true understanding but are very similar or add a distractor sentence to questions so that the model is guided to make incorrect inferences or answers.
            \item For ques-neg, rhetorical questions are preferred.
        \end{itemize}
    \textbf{STEP-3: Output JSON sample}
    \end{tcolorbox}
    \caption{Instructions for generating questions and answers based on extracted image information. GPT4 prompt for \textit{version 3}.}
    \label{fig:prompt_version3}
\end{figure*}

\begin{table*}
    \centering
\caption{Ablation about different combinations of data. "Version 2/3-Pos." denotes positive samples generated by the respective prompt version, while "Version 2/3-Neg." represents the corresponding negative samples.}
\vspace{0.3em}
\renewcommand{\arraystretch}{1.5}
\resizebox{.7\textwidth}{!}{\begin{tabular}{lccccc}
\specialrule{1.5pt}{0pt}{0pt}
\rowcolor[HTML]{EFEFEF}
{Dataset} & {MME$^\text{P}$} & {MME$^\text{C}$} &  MMB$^\text{T}$ & MMB$^\text{D}$ & POPE \\
\midrule
Version 2-Pos. & 714.63 & 241.07 & 59.52 & 60.65 & 74.85 \\
Version 2-Neg. & 500.0 & 200.0  & - & - & - \\
Version 2-Pos.+Neg. & 1289.56 & 276.43 & 59.70 & 61.43 & 75.78 \\
\midrule
Version 3-Pos. & 648.91 & 233.57 & 60.81 & 61.25 & 70.16 \\
Version 3-Neg. & 604.58 & 200.0 & 59.53 & 60.05 & 54.47 \\
Version 3-Pos.+Neg. & 1321.89 & 255.36 & 63.11 & 63.32 & 79.62 \\
\specialrule{1.5pt}{0pt}{0pt}
\end{tabular}}
    \label{tab:appendix_promptv2_3}
\end{table*}
\begin{table*}
    \centering
    \caption{Ablation about different combinations of data on MMVU test set. "Version 2/3-Pos." denotes positive samples generated by the respective prompt version, while "Version 2/3-Neg." represents the corresponding negative samples.}
    \vspace{0.3em}
    \setlength{\tabcolsep}{3pt}
    \renewcommand{\arraystretch}{1.5}
    \resizebox{\textwidth}{!}{%
        \begin{tabular}{lccccccccccccc}
            \specialrule{1.5pt}{0pt}{0pt}
            \rowcolor[HTML]{EFEFEF}
             Method & Char/Num & Pres. & Color/Tex & Num. & Shape & Stance & Pos. & Abstract. & Concrete. & Expert. & Act. & \multicolumn{1}{c}{Rel.} & Avg. RA$\uparrow$\\
            \hline
            Version 2-Pos. & 22.5 & 18.18 & 20.83 & 33.33 & 70.83 & 20.83 & 18.18 & 50.0 & 28.57 & 19.35 & 50.0 & 18.18 & 30.33\\
            Version 2-Neg. & 0.00 & 0.00 & 0.00 & 0.00 & 0.00 & 0.00 & 0.00 & 0.00 & 0.00 & 0.00 & 0.00 & 0.00 & 0.00\\
            Version 2-Pos.+Neg. & 20.0 & 9.09 & 20.83 & 33.33 & 58.33 & 25.0 & 13.64 & 45.45 & 23.81 & 19.35 & 50.0 & 18.18 & 27.67\\
            \hline
            Version 3-Pos. & 20.0 & 18.18 & 16.67 & 16.67 & 58.33 & 20.83 & 18.18 & 45.45 & 23.81 & 19.35 & 50.0 & 27.27 & 27.33\\
            Version 3-Neg. & 27.5 & 9.09 & 16.67 & 8.33 & 45.83 & 20.83 & 0.0 & 45.45 & 23.81 & 32.26 & 45.83 & 31.82 & 26.0\\
            Version 3-Pos.+Neg. & 40.0 & 18.18 & 50.0 & 33.33 & 70.83 & 25.0 & 22.73 & 63.64 & 57.14 & 41.94 & 58.33 & 40.91 & 43.33\\
            \specialrule{1.5pt}{0pt}{0pt}
            \rowcolor[HTML]{EFEFEF}
             Method & Char/Num & Pres. & Color/Tex & Num. & Shape & Stance & Pos. & Abstract. & Concrete. & Expert. & Act. & \multicolumn{1}{c}{Rel.} & Avg. MR $\downarrow$\\
            \hline
            Version 2-Pos. & 67.86 & 73.33 & 73.68 & 46.67 & 19.05 & 77.27 & 69.23 & 50.0 & 68.42 & 76.0 & 36.84 & 73.33 & 60.94\\
            Version 2-Neg. & 100.0 & 100.0 & 100.0 & 0 & 100.0 & 100.0 & 0 & 100.0 & 100.0 & 100.0 & 100.0 & 100.0 & 100.0\\
            Version 2-Pos.+Neg. & 72.41 & 80.0 & 70.59 & 50.0 & 36.36 & 71.43 & 66.67 & 52.38 & 73.68 & 76.0 & 42.86 & 75.0 & 63.27\\
            \hline
            Version 3-Pos. & 70.37 & 69.23 & 75.0 & 71.43 & 33.33 & 77.27 & 66.67 & 52.38 & 75.0 & 75.0 & 40.0 & 62.5 & 63.72\\
            Version 3-Neg. & 50.0 & 75.0 & 69.23 & 80.0 & 38.89 & 66.67 & 100.0 & 52.38 & 68.75 & 52.38 & 35.29 & 36.36 & 56.18\\
            Version 3-Pos.+Neg. & 44.83 & 66.67 & 25.0 & 38.46 & 22.73 & 70.0 & 44.44 & 33.33 & 33.33 & 45.83 & 30.0 & 43.75 & 40.91\\
            \specialrule{1.5pt}{0pt}{0pt}
        \end{tabular}
    }
    \label{tab:appendix_promptv2_3_mma}
\end{table*}

\section{Training an MLLM on the Proposed MMVU-Train}
\subsection{Framework.} The structure consists of a visual encoder, a visual-language connector, and a language model. Specifically, we employ SigLIP \cite{zhai2023sigmoid} as the visual encoder, Phi-2 (2.7B) \cite{microsoft2024phi2} as the language model, and a two-layer MLP as the connector. We build the model based on Bunny~\cite{he2024efficient}.

\subsection{Training Scheme.} Given an image and text, the image is processed through a visual encoder to obtain visual features. These features are then adjusted via a connector to align with the dimensions of the language model. The text is tokenized to generate textual features. These visual and textual features are concatenated and fed into the language model for generating responses. During training, each sample comprises instruction and response. The instructions are masked, and only the response and the model's output are used to calculate the loss.

We employ the two-stage training strategy. In the first stage, the visual and language alignment stage, we use \textit{Bunny-pretrain-LAION-2M} \cite{he2024efficient} data to train only the connector while freezing the vision encoder and the language model. In the second stage, we fine-tune both the connector and the language model using the generated MMVU dataset (MMVU-Train). To further enrich the dataset's diversity, we include OCR data (OCR-VQA \cite{mishra2019ocr}), visual question-answering data (VQA-v2 \cite{goyal2017making}, GQA \cite{GQA2019}, OK-VQA \cite{marino2019ok}, A-OKVQA \cite{schwenk2022okvqa}, Visual Genome \cite{krishna2017visual}, and RefCOCO \cite{kazemzadeh2014referitgame}). Additionally, to maintain the language capabilities of the MLLM, we incorporated text-only data from WizardLM \cite{xu2023wizardlm}.

\subsection{Implementation Details.} In the visual and language alignment stage, we train only the connector for one epoch with a learning rate of 1e-5. In the fine-tuning phase, we fine-tune both the connector and the language model using LoRA~\cite{hu2021lora} with a learning rate of 2e-4. The training is implemented in PyTorch using 8 Nvidia A100 GPUs in an internal server.

\subsection{Evaluation Benchmark.} We evaluate the performance of different MLLMs with the MMVU test set (MMVU$^\text{A}$ and MMVU$^\text{M}$ denote the average score of RA and MR, respectively.), and general benchmarks. We utilize commonly used general benchmarks: MME perception (MME$^\text{P}$) \cite{mme}, MME cognition (MME$^\text{C}$) \cite{mme}, MMBench test split (MMB$^\text{T}$) \cite{mmbench}, MMBench dev split (MMB$^\text{D}$) \cite{mmbench}, SEED-Bench-1 (SEED) \cite{seedbench}, MMMU validation split (MMMU$^\text{V}$) \cite{yue2023mmmu}, MMMU test split (MMMU$^\text{T}$) \cite{yue2023mmmu}, VQA-v2 test-dev split \cite{goyal2017making}, GQA test-dev-balanced split \cite{GQA2019}, and the average F1-score across random, popular, and adversarial categories on the validation set of MSCOCO (POPE) \cite{li-etal-2023-evaluating}. This comprehensive validation ensures robust evaluation across diverse metrics and scenarios.

\subsection{Expriment Results}
\paragraph{Comparison of MLLMs.} We augment our generated MMVU-Train with additional datasets to form a comprehensive new dataset. We validate this expanded dataset on the proposed MMVU and several general benchmarks. As shown in Tab.~\ref{tab:main}, Experiments show that the 3-billion parameter model trained on this enriched dataset outperforms the LLaVA-v1.5-7B on most benchmarks. Remarkably, it also exceeds the performance of some models with 13-billion parameters on several benchmarks.

\begin{table*}
    \centering
    \caption{{ Comparison to leading MLLMs on 10 benchmarks.}
    $^\mathsection$We evaluate the officially released checkpoint by ourselves.}
    \vspace{0.3em}
   \setlength{\tabcolsep}{3pt}
 \resizebox{\textwidth}{!}{%
        \begin{tabular}{lllccccccccccc}
            \specialrule{1.5pt}{0pt}{0pt}
            \rowcolor[HTML]{EFEFEF}
            Model & Vision Encoder & LLM & MME$^\text{P}$ & MME$^\text{C}$ & MMB$^\text{T}$ & MMB$^\text{D}$ & SEED & MMMU$^\text{V}$ & MMMU$^\text{T}$ & VQA$^\text{v2}$ & GQA & POPE \\
            \hline
            IDEFICS-80B \cite{idefics} & OpenCLIP-H (1.0B) & LLaMA-65B & -- & -- & 54.6 & 54.5 & -- & -- & -- & 60.0 & -- & -- \\
            BLIP-2 \cite{li2023blip} & EVA01-CLIP-G (1.0B) & Vicuna-13B & -- & -- & -- & -- & -- & -- & -- & -- & 41.0 & -- \\
            InstructBLIP \cite{instructblip} & EVA01-CLIP-G (1.0B) & Vicuna-13B & -- & -- & -- & -- & -- & -- & -- & -- & 49.5 & 83.7 \\
            BLIP-2 \cite{li2023blip} & EVA01-CLIP-G (1.0B) & Flan-T5-XXL (11B) & 1293.8 & 290.0 & -- & -- & -- & 35.4 & 34.0 & 65.0 & 44.6 & -- \\
            InstructBLIP \cite{instructblip} & EVA01-CLIP-G (1.0B) & Flan-T5-XXL (11B) & 1212.8 & 291.8 & -- & -- & -- & 35.7 & 33.8 & -- & 47.9 & -- \\
            Shikra-13B \cite{chen2023shikra} & CLIP-L (0.4B) & Vicuna-13B & -- & -- & -- & -- & -- & -- & -- & 77.4 & -- & -- \\
            LLaVA-v1.5-13B (LoRA) \cite{liu2023improvedllava} & CLIP-L (0.4B) & Vicuna-13B & 1541.7 & 300.4$^\mathsection$ & 68.4$^\mathsection$ & 68.5 & 61.3 & 40.0$^\mathsection$ & 33.2$^\mathsection$ & 80.0 & 63.3 & 86.7 \\
            VILA-13B~\cite{lin2023vila} & CLIP-L (0.4B) & LLaMA2-13B & 1570.1 & - & 70.3 & - & 62.8 & - & - & 80.8 & 63.3 & 84.2 \\
            SPHINX-Plus~\cite{gao2024sphinx} & Mixture of Visual Experts (MoV) &  LLaMA2-13B & 1457.7 & 283.6 & 71.0 & - & 74.8 & - & - & - & - & 89.1 \\
            LLaVA-Llama-3-8B-v1.1~\cite{Xtuner} & CLIP-L (0.4B) & Llama-3-8B & 1469 & 349 & 72.3 & - & - & 36.8 & - & - & 62.6 & 86.4\\
            InstructBLIP \cite{instructblip} & EVA01-CLIP-G (1.0B) & Vicuna-7B & -- & -- & 33.9 & 36.0 & 53.4 & -- & -- & -- & 49.2 & -- \\
            MiniGPT-v2 \cite{chen2023minigptv2} & EVA01-CLIP-G (1.0B) & LLaMA2-7B & -- & -- & -- & -- & -- & -- & -- & -- & 60.3 & -- \\
            IDEFICS-9B \cite{idefics} & OpenCLIP-H (1.0B) & LLaMA-7B & -- & -- & 45.3 & 48.2 & -- & -- & -- & 50.9 & -- & -- \\
            LLaVA-v1.5-7B (LoRA) \cite{liu2023improvedllava} & CLIP-L (0.4B) & Vicuna-7B & 1476.9 & 267.9$^\mathsection$ &  66.1$^\mathsection$ & 66.1 & 60.1 & 34.4$^\mathsection$ & 31.7$^\mathsection$ & 79.1 & 63.0 & 86.4 \\
            mPLUG-Owl2 \cite{ye2023mplugowl2} & CLIP-L (0.4B) & LLaMA2-7B & 1450.2 & 313.2 & 66.0 & 66.5 & 57.8 & 32.7 & 32.1 & 79.4 & 56.1 & 85.8 \\
            VILA-7B~\cite{lin2023vila} & CLIP-L (0.4B) & LLaMA2-7B & 1533.0 & - & 68.9 & - & 61.1 & - & - & 79.9 & 62.3 & 85.5 \\
            Shikra-7B \cite{chen2023shikra} & CLIP-L (0.4B)& Vicuna-7B & -- & -- & 60.2 & 58.8 & -- & -- & -- & -- & -- & -- \\
            SPHINX-Intern2~\cite{gao2024sphinx} & Mixture of Visual Experts (MoV) & InternLM2-7B & 1260.4 & 294.6 & 57.9 & - & 68.8 & - & - & 75.5 & 56.2 & 86.9\\
            Mini-Gemini~\cite{li2024mini} & CLIP & Vicuna-7B & 1523 & 316 & - & - & - & 36.1 & 32.8 & 65.2 & - & -\\
            MM1-7B-Chat~\cite{mckinzie2024mm1} & ViT-H (5B) & MM1-7B & 1529.3 & 328.9 & 72.3 & - & 64.0 & 37.0 & 35.6 & 82.8 & - & 86.6 \\
            LLaVA-Phi-3-mini~\cite{Xtuner} & CLIP-L (0.4B) & Phi-3 (4B) & 1477 & 313 & 69.2 & - & - & 41.4 & - & - & 61.5 & 87.3\\
            Gemini Nano-2 \cite{team2023gemini} & -- & Gemini Nano-2 (3.25B) & -- & -- & -- & -- & -- & 32.6 & -- & 67.5 & -- & -- \\
            MM1-3B-Chat~\cite{mckinzie2024mm1} & ViT-H (5B) & MM1-3B & 1482.5 & 279.3 & 67.8 & - & 63.0 & 33.9 & 33.7 & 82.0 & - & 87.4\\
            MobileVLM \cite{chu2023mobilevlm} & CLIP-L (0.4B) & MobileLLaMA (2.7B) & 1288.9 & -- & -- & 59.6 & -- & -- & -- & -- & 59.0 & 84.9 \\
            MobileVLM V2 \cite{chu2024mobilevlm} & CLIP-L (0.4B) & MobileLLaMA (2.7B) & 1440.5 & -- & -- & 63.2 & -- & -- & -- & -- & 61.1 & 84.7 \\
            ALLaVA-Longer~\cite{chen2024allava} & SigLIP-SO (0.4B) & Phi-2 (2.7B) & - & - & 64.6 & - & 65.6 & 33.2 & - & - & 50.0 & - \\
            TinyLLaVA-share-Sig-Phi~\cite{zhou2024tinyllava} & SigLIP-SO (0.4B) & Phi-2 (2.7B) & 1464.9 & - & 66.9 & - & - & - & - & 79.9 & 62.0 & 86.4\\
            \midrule
            \textbf{Bunny-MMVU-3B} & SigLIP-SO (0.4B) & Phi-2 (2.7B) & 1515.1 & 262.9 & 70.7 & 69.7 & 63.2 & 38.2 & 34.0& 80.1 & 62.1 & 86.0\\
            \textbf{Bunny-MMVU-4B} & SigLIP-SO (0.4B) & Phi-3 (4B) & 1511.2 & 313.2 & 72.3 & 71.8 & 64.1 & 40.9 & 38.8 & 80.4 & 62.2 & 86.3\\
            \textbf{Bunny-MMVU-8B} & SigLIP-SO (0.4B) & Llama-3-8B & 1606.1 & 331.8 & 77.6 & 76.3 & 65.1 & 42.6 & 38.7 & 81.5 & 63.4 & 86.3\\
            \specialrule{1.5pt}{0pt}{0pt}
        \end{tabular}
    }
    \label{tab:main}
\end{table*}

\paragraph{Ablation about combining the proposed MMVU-Train with other datasets.} We conduct experiments using an expanded dataset to evaluate the impact of integrating the MMVU-Train with other datasets. We sample 24k instances from the LLaVA 158k dataset and combine them with other datasets to construct 106k samples (Mix 106k), maintaining proportional representation. We explore various combination methods, such as replacing the original LLaVA data or directly adding MMVU data. Our results, presented in Tab.~\ref{tab:ablation_combination_with_other_dataset.tex}, reveal that adding new data does not consistently improve performance; replacement yields more substantial enhancements. The model achieves optimal performance when the replacement proportion closely matches the original distribution.

\begin{table*}
    \centering
\caption{Ablation about the combination of the proposed MMVU dataset with other datasets. We train these datasets with the same vision encoder (SigLIP).}
\label{tab:ablation_combination_with_other_dataset.tex}
\vspace{0.3em}
\resizebox{\textwidth}{!}{\begin{tabular}{lccccccccccccc}
\specialrule{1.5pt}{0pt}{0pt}
\rowcolor[HTML]{EFEFEF}
{Dataset} & LLM & MMVU$^\text{A}$ $\uparrow$ & MMVU$^\text{M}$ $\downarrow$ & MME$^\text{P}$ & MME$^\text{C}$ & MMB$^\text{T}$ & MMB$^\text{D}$ & SEED & MMMU$^\text{V}$ & MMMU$^\text{T}$ & VQA$^\text{v2}$ & GQA & POPE \\
\hline
Mix 106k & Phi-2 & 44.00 & 43.59 & 1410.01 & 266.07 & 66.14 & 66.15 & 60.07 &  36.70 & 32.90 & 75.71 & 56.39 & 84.89\\
Mix 106k-replace MMVU-Train 12k & Phi-2 & 45.33 & 41.63 & 1432.51 & 271.78 & 61.43 & 60.91 & 57.97 & 35.00 & 33.20 & 75.75 & 56.50 & 83.64\\
Mix 106k-replace MMVU-Train 24k & Phi-2 & 47.00 & 40.51 & 1421.54 & 248.57 & 67.26 & 66.32 & 60.37 & 38.00 & 34.00 & 75.95 & 56.92 & 84.40\\
Mix 106k-replace MMVU-Train 48k & Phi-2 & 47.33 & 38.26 & 1419.26 & 239.29 & 66.76 & 65.64 & 60.61 & 35.60 & 33.70 & 76.24 & 56.96 & 83.63\\
Mix 106k + MMVU-Train 12k & Phi-2 & 50.00 & 36.71 & 1388.89 & 252.5 & 65.86 & 66.07 & 60.51 & 37.30 & 33.00 & 75.86 & 56.87 & 84.06\\
Mix 106k + MMVU-Train 24k & Phi-2 & 48.67 & 38.66 & 1400.74 & 244.64 & 66.98 & 65.98 & 60.62 & 37.10 & 32.90 & 76.05 & 56.96 & 83.73\\
Mix 106k + MMVU-Train 48k & Phi-2 & 47.00 & 38.16& 1410.40 & 247.5 & 65.86 & 66.24 & 60.85 & 36.80 & 33.90 & 73.45 & 57.05 & 84.98\\
\midrule
Base 695k & Phi-2 & 55.33 & 27.19 & 1492.17 & 274.29 & 68.27 & 68.56 & 62.31 & 37.40 & 32.80 & 79.85 & 61.84 & 86.52\\
Base-replace MMVU-Train 112k (649k) & Phi-2 & 57.00 & 22.97 & 1498.57 & 255.35 & 70.35 & 68.90 & 62.88 & 35.70 & 33.40 & 79.69 & 62.08 & 85.13\\
Weight merge & Phi-2 & 58.33 & 23.91 & 1515.06 & 262.86 & 70.74 & 69.67 & 63.23 & 38.20 & 34.00 & 80.14 & 62.10 & 85.98\\
\midrule
Base 695k & Phi-3 & 54.67 & 29.61 & 1462.07 & 335.35 & 70.57 & 71.56 & 63.08 & 40.30 & 39.00 & 80.34 & 61.69 & 85.21\\
Base-replace MMVU-Train 112k (649k)  & Phi-3 & 58.33 & 24.57 & 1520.43 & 283.93 & 72.43 & 71.23& 63.89 & 41.30 & 38.50 & 80.14 & 61.74 & 85.88\\
Weight merge & Phi-3 & 58.33 & 27.39 & 1511.16 & 313.2 & 72.31 & 71.82 & 64.12 & 40.90 & 38.80 & 80.44 & 62.21 & 86.28\\
\midrule
Base 695k & Llama-3-8B & 58.00 & 26.27 & 1548.53 & 313.57 & 76.63& 75.34 & 64.62 & 40.70 & 38.60 & 80.99 & 62.82 & 85.30\\
Base-replace MMVU-Train 112k (649k)  & Llama-3-8B & 61.00 & 21.12 & 1600.46 & 318.93 & 75.22& 75.26 & 64.62 & 41.70 & 37.90 & 81.03 & 62.72 & 86.30\\
Weight merge & Llama-3-8B & 59.33 & 24.58 & 1606.12 & 331.79 & 77.63 & 76.29 & 65.11 & 42.60 & 38.70 & 81.52 & 63.44 & 86.25\\
\specialrule{1.5pt}{0pt}{0pt} 
\end{tabular}}
\end{table*}

\section{Examples of Paired Positive and Negative Samples Generated}
\begin{figure*}[h!]
    \centering
    \begin{tcolorbox}[colback=white, colframe=black, before=\par\bigskip]
    \centering
    \includegraphics[width=0.6\textwidth]{images/generation_excample.png}
    \label{fig:kitchen}
            \begin{tcolorbox}[colback=white, colframe=gray, title=Conversation]
                \textbf{Question:} Are the people preparing for a party? \\
                \textbf{Answer:} No, they are preparing a meal. \\
                
                \textbf{Question:} Is there a green chair visible? \\
                \textbf{Answer:} No, no green chair. \\
                
                \textbf{Question:} Are there three people in the kitchen? \\
                \textbf{Answer:} No, only two people. \\
                
                \textbf{Question:} Is the kitchen light pink? \\
                \textbf{Answer:} No, it is not pink. \\
                
                \textbf{Question:} Does the window show a garden outside? \\
                \textbf{Answer:} No visible garden.
            \end{tcolorbox}
    \end{tcolorbox}
    \caption{Generated conversation example by prompt version 0.}
    \label{fig: generation_excample_v0}
\end{figure*}
\begin{figure*}[h!]
    \centering
    \begin{tcolorbox}[colback=white, colframe=black, before=\par\bigskip]
        \centering
        \begin{minipage}{0.9\textwidth}
            \begin{tcolorbox}[colback=white, colframe=gray, title=Conversation (Question type: Character)]
                \textbf{Question:} Is there a number 358 visible anywhere? \\
                \textbf{Answer:} No, there are no numbers visible. \\
                
                \textbf{Question:} Does the text 'Happy Birthday' appear in the kitchen? \\
                \textbf{Answer:} No, there is no text visible. \\
            \end{tcolorbox}
            
            \begin{tcolorbox}[colback=white, colframe=gray, title=Conversation (Question type: Semantic)]
                \textbf{Question:} Is there a microwave in the image? \\
                \textbf{Answer:} No, there is no microwave visible. \\
                
                \textbf{Question:} Can you see a refrigerator in the scene? \\
                \textbf{Answer:} No, there is no refrigerator visible. \\
                
                \textbf{Question:} Is there a blender on the counter? \\
                \textbf{Answer:} No, there is no blender shown. \\
                
                \textbf{Question:} Are there five plates in the picture? \\
                \textbf{Answer:} No, there are more than five plates visible. \\
            \end{tcolorbox}
            
            \begin{tcolorbox}[colback=white, colframe=gray, title=Conversation (Question type: Understanding)]
                \textbf{Question:} Are the individuals preparing for a meal? \\
                \textbf{Answer:} Yes, serving food. \\
                
                \textbf{Question:} Is the setting a restaurant kitchen? \\
                \textbf{Answer:} No, it's a home kitchen. \\
            \end{tcolorbox}
        \end{minipage}
    \end{tcolorbox}
    \caption{Generated conversation example by prompt version 1.}
    \label{fig: generation_excample_v1}
\end{figure*}
\begin{figure*}[h!]
    \centering
    \begin{tcolorbox}[colback=white, colframe=black, before=\par\bigskip]
    \centering
        \begin{minipage}{0.9\textwidth}
            \begin{tcolorbox}[colback=white, colframe=gray, title=Positive Conversation]
                \textbf{Question:} Is there a bowl of food on the counter? \\
                \textbf{Answer:} Yes, there are multiple bowls of food on the counter. \\
                
                \textbf{Question:} What is the man wearing? \\
                \textbf{Answer:} The man is wearing a plaid shirt. \\
                
                \textbf{Question:} How many people are in the kitchen? \\
                \textbf{Answer:} There are two people in the kitchen. \\
                
                \textbf{Question:} What are the people doing? \\
                \textbf{Answer:} They are preparing a meal. \\
                
                \textbf{Question:} Where is the sink located in the kitchen? \\
                \textbf{Answer:} The sink is located towards the right. \\
            \end{tcolorbox}
            
            \begin{tcolorbox}[colback=white, colframe=gray, title=Negative Conversation]
                \textbf{Question:} Is there a vase of flowers on the counter? \\
                \textbf{Answer:} No, there is no vase of flowers on the counter. \\
                
                \textbf{Question:} What is the man wearing? \\
                \textbf{Answer:} The man is wearing a green hoodie. \\
                
                \textbf{Question:} How many children are in the kitchen? \\
                \textbf{Answer:} There are no children in the kitchen. \\
                
                \textbf{Question:} What are some people doing? \\
                \textbf{Answer:} They are reading books. \\
                
                \textbf{Question:} Where is the dishwasher located in the kitchen? \\
                \textbf{Answer:} The dishwasher is located towards the right. \\
            \end{tcolorbox}
        \end{minipage}
    \end{tcolorbox}
    \caption{Generated conversation example by prompt version 2.}
    \label{fig: generation_excample_v2}
\end{figure*}

\begin{figure*}[h!]
    \centering
    \begin{tcolorbox}[colback=white, colframe=black, before=\par\bigskip]
        \centering
        \begin{minipage}{\textwidth}
            \begin{tcolorbox}[colback=white, colframe=gray, title=Positive Conversation]
                \textbf{Question:} Are there two people in the kitchen? \\
                \textbf{Option:} \{"Yes, there are two people.", "Yes, there are three people.", "No, there is only one person.", "No, there are four people."\} \\
                \textbf{Answer:} Yes, there are two people. \\
                
                \textbf{Question:} Are the people in the kitchen cooking? \\
                \textbf{Option:} \{"Yes, they are cooking.", "No, they are cleaning.", "No, they are eating.", "Yes, they are fixing something."\} \\
                \textbf{Answer:} Yes, they are cooking. \\
                
                \textbf{Question:} How many bowls are there on the counter? \\
                \textbf{Option:} \{"One bowl.", "Four bowls.", "Six bowls.", "Two bowls."\} \\
                \textbf{Answer:} Four bowls. \\
                
                \textbf{Question:} What is the main activity happening? \\
                \textbf{Option:} \{"Preparing a meal.", "Washing dishes.", "Playing games.", "Reading books."\} \\
                \textbf{Answer:} Preparing a meal. \\
                
                \textbf{Question:} Is there a dishwasher in the image? \\
                \textbf{Option:} \{"Yes, the dishwasher is there.", "No, there's no dishwasher.", "It's unclear.", "There is a microwave instead."\} \\
                \textbf{Answer:} Yes, the dishwasher is there. \\
                
                \textbf{Question:} Is there a pot on the table? \\
                \textbf{Option:} \{"Yes, there is a pot.", "No, there isn't a pot.", "Yes, there are two pots.", "No, there are only pans."\} \\
                \textbf{Answer:} Yes, there is a pot. \\
                
                \textbf{Question:} Are the people preparing food in a modern kitchen? \\
                \textbf{Option:} \{"Yes, they are.", "No, it's an old kitchen.", "Yes, in a living room.", "No, it's a commercial kitchen."\} \\
                \textbf{Answer:} Yes, they are. \\
            \end{tcolorbox}
        \end{minipage}
    \end{tcolorbox}
    \caption{Generated conversation example by prompt version 3 (part 1).}
    \label{fig: generation_excample_v3_p1}
\end{figure*}
\begin{figure*}[h!]
    \centering
    \begin{tcolorbox}[colback=white, colframe=black, before=\par\bigskip]
        \centering
        \begin{minipage}{\textwidth}
            \begin{tcolorbox}[colback=white, colframe=gray, title=Negative Conversation]
                \textbf{Question:} Are there three people in the kitchen? \\
                \textbf{Option:} \{"Yes, there are two people.", "Yes, there are three people.", "No, only one person.", "No, four people."\} \\
                \textbf{Answer:} No, only one person. \\
                
                \textbf{Question:} Are the people in the kitchen cleaning? \\
                \textbf{Option:} \{"Yes, they are cooking.", "No, they are cleaning.", "No, they are eating.", "Yes, fixing something."\} \\
                \textbf{Answer:} Yes, they are cooking. \\
                
                \textbf{Question:} How many bowls are there on the countertop? \\
                \textbf{Option:} \{"One bowl.", "Four bowls.", "Six bowls.", "Two bowls."\} \\
                \textbf{Answer:} Four bowls. \\
                
                \textbf{Question:} What is the secondary activity happening? \\
                \textbf{Option:} \{"Preparing a meal.", "Washing dishes.", "Playing games.", "Reading books."\} \\
                \textbf{Answer:} Preparing a meal. \\
                
                \textbf{Question:} Is there a microwave in the image? \\
                \textbf{Option:} \{"Yes, the microwave is there.", "No, there's no microwave.", "It's unclear.", "There is a dishwasher."\} \\
                \textbf{Answer:} There is a dishwasher. \\
                
                \textbf{Question:} Is there a pan on the table? \\
                \textbf{Option:} \{"Yes, there is a pot.", "No, there isn't a pot.", "Yes, there are two pans.", "No, only bowls."\} \\
                \textbf{Answer:} Yes, there is a pot. \\
                
                \textbf{Question:} Are the people preparing food in a commercial kitchen? \\
                \textbf{Option:} \{"Yes, they are.", "No, it's old kitchen.", "Yes, a living room.", "No, modern kitchen."\} \\
                \textbf{Answer:} No, modern kitchen. \\
            \end{tcolorbox}
        \end{minipage}
    \end{tcolorbox}
    \caption{Generated conversation example by prompt version 3 (part 2).}
    \label{fig: generation_excample_v3_p2}
\end{figure*}

\section{The Details of ``Instruction prompt''.}
\begin{figure*}[h!]
    \centering
    \begin{tcolorbox}[width=\textwidth, colframe=black!75, colback=white!95!gray, sharp corners=all]
        \small
        \textbf{STEP-1 (DO NOT OUTPUT): Image Information Extraction}
        \begin{enumerate}[leftmargin=10mm]
            \item \textbf{Text/Numbers:} Identify clear text or numbers in the image and their positions.
            \item \textbf{Objects:} Identify objects, their properties (name/category/color/texture/shape/pose), position, and quantity.
            \item \textbf{People:} Identify people, and their attributes (name, position, gender, facial expression), and distinguish between real people and portraits in photos or posters.
            \item \textbf{Object Relationships:} Understand relationships between objects (relative positions or hierarchy).
            \item \textbf{People-Object Relationships:} Understand relationships between people and objects (e.g., someone riding a bike).
            \item \textbf{Activities:} If there are multiple people, identify their activities.
            \item \textbf{Events:} Understand events in the image and their cultural, emotional, or knowledge-related context.
        \end{enumerate}
        \vspace{4mm}
        \textbf{STEP-2 (DO NOT OUTPUT): Carefully review the image and the question to ensure an accurate understanding of each question.}
        \vspace{4mm}
        \textbf{Example Responses:}
        \begin{itemize}[leftmargin=10mm]
            \item \textbf{Question:} Is there a red bicycle in the image?\\
            \textbf{Answer:} Yes, there is a red bicycle in the image. It is clearly shown in the center.
            \item \textbf{Question:} Is the car in the image green?\\
            \textbf{Answer:} No, the car in the image is blue. The color is distinct.
        \end{itemize}
        \vspace{4mm}
        \textbf{STEP-3: Answer with the option's letter from the given choices directly.}
    \end{tcolorbox}
    \caption{Common Prompt for information extraction.}
    \label{fig:prompt_common}
\end{figure*}

\section{Details about Content Guided Refinement Strategy}

\subsection{Prompt for extraction visual information}
The prompt is shown in Fig.~\ref{fig:Prompt for information extraction}.
\begin{figure*}[h!]
    \centering
    \begin{tcolorbox}[width=\textwidth, colframe=black!75, colback=white!95!gray, sharp corners=all]
        \small
        \textbf{Extract information from the image and structure the output as follows:}
        \begin{enumerate}[leftmargin=10mm]
            \setlength{\itemsep}{4pt}
            \item \textbf{Text/Numbers:} Identify any visible text or numbers, including their content, font, and precise positions in the image.
            \item \textbf{Objects:} Identify each object, detailing its name/category, color, texture, shape, pose, relative location within the image, and quantity.
            \item \textbf{People:} Identify all visible individuals, with attributes such as position, gender, facial expression, and distinguish between real individuals and representations (e.g., in portraits, posters).
            \item \textbf{Relationships \& Interactions:} Describe spatial or contextual relationships between objects, and interactions or spatial relationships between people and objects.
            \item \textbf{Activities:} Identify and describe any observed activities or actions involving people or objects.
            \item \textbf{Events:} Identify any significant events or scenes occurring in the image.
        \end{enumerate}
    \end{tcolorbox}
    \caption{Prompt for information extraction.}
    \label{fig:Prompt for information extraction}
\end{figure*}
\subsection{Examples of Content Guided Refinement Strategy Results}
\begin{figure*}[h]
    \centering
    \begin{tcolorbox}[colback=white, colframe=black, before=\par\bigskip]
        \centering
        \includegraphics[width=0.6\textwidth]{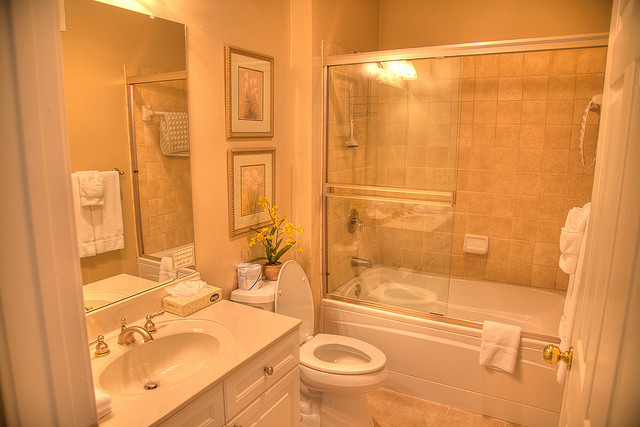}
        \label{fig:kitchen}
        \begin{tcolorbox}[colback=white, colframe=gray, title=Example]
            \small
            \textbf{ID:} attribute/existence/COCO\_val2014\_000000170365\\
            \textbf{Image:} COCO\_val2014\_000000170365.jpg\\
            \textbf{Question:} What is the object under a plate of yellow plants in the picture?\\
            \textbf{Options:} \\
            A. Towel \\
            B. Shower enclosure \\
            C. Sink \\
            D. Closestool \\
            \textbf{Ground Truth:} D \\
            \textbf{Question Type:} POS \\
            \textbf{Info:} \\
            \begin{enumerate}
                \item \textbf{Text/Numbers:} There is no visible text or numbers in the image.
                \item \textbf{Objects:}
                \begin{enumerate}
                    \item Sink: A white porcelain sink with a silver faucet.
                    \item Toilet: A white porcelain toilet.
                    \item Bathtub: A white porcelain bathtub with a shower door.
                    \item Mirror: A framed mirror on the wall above the sink.
                    \item Potted Plant: A yellow plant in a pot on the counter.
                    \item Towels: White towels hanging on the towel rack.
                    \item Toilet Paper: A roll of toilet paper on the back of the toilet.
                    \item Tissue Box: A box of tissues on the counter.
                    \item Trash Can: A white trash can with a lid.
                    \item Toilet Brush: A white toilet brush in a holder.
                \end{enumerate}
                \item \textbf{Relationships \& Interactions:} Various spatial and functional relationships between objects, as described above.
                \item \textbf{Activities:} None observed.
                \item \textbf{Events:} The image shows a typical bathroom setup.
            \end{enumerate}
        \end{tcolorbox}
    \end{tcolorbox}
    \caption{Examples of Content Guided Refinement Strategy Results.}
    \label{fig:CRG_example}
\end{figure*}

The example is shown in Fig.~\ref{fig:CRG_example}.

\section{Details about Visual Attention Refinement Strategy}

\subsection{Detailed Description}
Our analysis indicates that the probability of the model generating correct answers decreases when the attention values between the question token and the visual token are relatively low. To address this, we propose leveraging the attention mechanism between the question and visual tokens to optimize the visual prompts, as shown in the algorithm ~\ref{Optimizing Visual Prompts Using Attention Mechanisms}. Specifically, we extract the attention parameters from the final transformer layer of the MLLM, while other proposed steps remain consistent with those outlined in the main text. Using these parameters, an attention map is generated, where regions with higher values indicate areas of focus for the model. To guide the model toward regions relevant to the question, we invert the attention map, apply regularization and filtering, and blend it with the original image. The blending process employs coefficients of 0.85 for the original image and 0.15 for the attention map. 

\begin{algorithm}
\caption{Optimizing Visual Prompts Using Attention Mechanisms}
\label{alg:optimize_visual_prompt}
\begin{algorithmic}[1]
\REQUIRE MLLM with pre-trained parameters, input question $Q$, input image $I$, blending coefficients $\alpha = 0.85$, $\beta = 0.15$
\ENSURE Optimized visual prompt $I'$ for inference

\STATE Extract attention parameters $A$ from the final transformer layer of the MLLM.
\STATE Compute the attention map $M$ using $A$ between the question token and the visual tokens.
\STATE Identify high-value regions in $M$ where the model focuses during inference.

\STATE \textbf{Invert the attention map:}
\STATE Compute the inverted attention map $M_{\text{inv}} = 1 - M$.

\STATE \textbf{Regularize and filter:}
\STATE Apply normalization and filtering to $M_{\text{inv}}$ to obtain $M_{\text{filtered}}$.

\STATE \textbf{Blend attention map with the original image:}
\STATE Generate the optimized image $I'$ as:
\[
I' = \alpha \cdot I + \beta \cdot M_{\text{filtered}}
\]

\RETURN Optimized image $I'$ for inference.
\end{algorithmic}
\label{Optimizing Visual Prompts Using Attention Mechanisms}
\end{algorithm}

\subsection{Examples of visual attention refinement strategy}
\begin{figure*}[h!]
    \centering
    \begin{tcolorbox}[colback=white, colframe=black, before=\par\bigskip]
        \centering
        \includegraphics[width=0.4\textwidth]{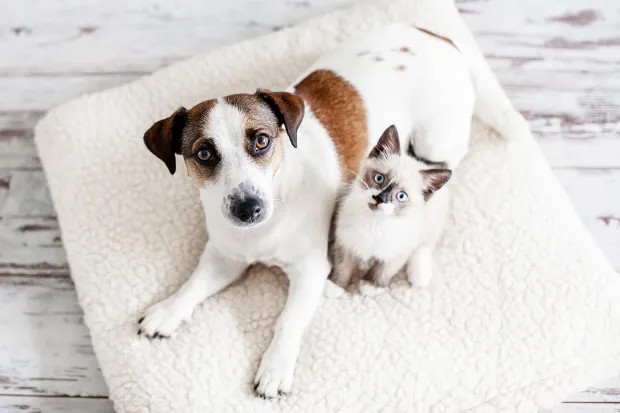}
        \includegraphics[width=0.4\textwidth]{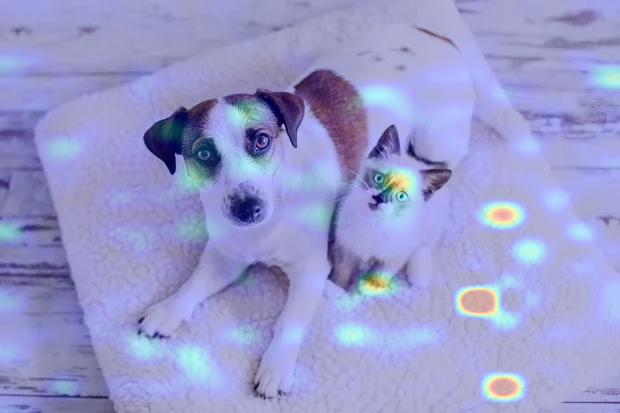}
        \label{fig:puppy_cat_1}
        \begin{tcolorbox}[colback=white, colframe=gray, title= POS Example]
            \small
            \textbf{ID:} attribute/existence/21\\
            \textbf{Image:} 21.jpg\\
            \textbf{Question:} How many puppies are there in the picture?\\
            \textbf{Options:} \\
            A. There is only one puppy in the picture and the other one is a cat.\\
            B. There are two puppies in the picture.\\
            C. There are no puppies in the picture, just two kittens.\\
            D. There is no puppy in the picture, just two stuffed animals.\\
            \textbf{Ground Truth:} A\\
            \textbf{Question Type:} POS\\
            \textbf{Question:} Does the picture show a dog and its pups?\\
            \textbf{Options:} \\
            A. The pup lay on top of the dog.\\
            B. The picture shows a dog and its two puppies.\\
            C. There are no puppies in the picture, just two kittens.\\
            D. There is a dog in the picture and the other smaller animal is a cat.\\
            \textbf{Ground Truth:} D\\
            \textbf{Question Type:} NEG\\
            \textbf{Info:}\\
            \begin{enumerate}
                \item \textbf{Text/Numbers:} There is no visible text or numbers in the image.
                \item \textbf{Objects:}
                \begin{enumerate}
                    \item Dog: White and brown, smooth fur, oval body, sitting on a blanket.
                    \item Cat: White and gray, soft fur, pointed ears, sitting on a blanket.
                    \item Blanket: White, soft, fluffy, laying on a wooden surface.
                    \item Wooden Surface: White, rough and uneven, under the blanket.
                \end{enumerate}
                \item \textbf{People:} No visible people.
                \item \textbf{Identification:} No real individuals, likely a portrait or poster.
                \item \textbf{Activities:} No observed actions.
                \item \textbf{Events:} No significant events.
            \end{enumerate}
        \end{tcolorbox}
    \end{tcolorbox}
    \caption{Examples of visual attention refinement strategy.}
    \label{Examples of visual attention refinement strategy}
\end{figure*}
The example is shown in Fig.~\ref{Examples of visual attention refinement strategy}.

\section{Broader Impact and Limitation}
\paragraph{Broader impact.} In this paper, we study how to evaluate and improve MLLMs' robustness when answering challenging visual questions. The research will benefit the general development and applications of MLLMs.

\paragraph{Limitation.} Due to the cost of dataset construction, we contribute 112K instruction tuning data. Such adversarial datasets can be scaled up with more quota.

\clearpage

\end{document}